\documentclass[reqno, table]{amsart}
\usepackage[english]{babel}
\usepackage{lipsum}
\makeatletter
\usepackage{amsfonts}
\usepackage{amssymb}
\usepackage{centernot}
\usepackage{graphicx}
\usepackage{algorithm}
\usepackage{algorithmic}
\usepackage[algo2e]{algorithm2e}
\usepackage{etoolbox} 
\usepackage{fullpage}
\usepackage[mathlines]{lineno}
\usepackage{tikz}
\usepackage{amsmath, amssymb}
\usetikzlibrary{arrows,positioning,shapes,decorations.pathreplacing, calc, arrows.meta}
\usepackage[super]{nth}
\usepackage{url}
\usepackage{hyperref}
\usepackage{xcolor}
\usepackage{multirow}
\usepackage[normalem]{ulem}
\usepackage{caption}
\usepackage{subcaption}
\newcommand{\authorfootnotes}{\renewcommand\thefootnote{\@fnsymbol\c@footnote}}%
\makeatother
\newcommand*\linenomathpatch[1]{%
  \cspreto{#1}{\linenomath}%
  \cspreto{#1*}{\linenomath}%
  \csappto{end#1}{\endlinenomath}%
  \csappto{end#1*}{\endlinenomath}%
}

\linenomathpatch{equation}
\linenomathpatch{gather}
\linenomathpatch{multline}
\linenomathpatch{align}
\linenomathpatch{alignat}
\linenomathpatch{flalign}

\colorlet{red-}{black}
\colorlet{blue-}{black}
\colorlet{orange-}{black}

\DeclareMathOperator*{\argmin}{arg\,min}
\DeclareMathOperator{\sgn}{sgn}

\newcommand{\bb}[1]{\mathbf{#1}}

\definecolor{darkgreen}{rgb}{.15,.55,0}

\definecolor{darkblue}{rgb}{0,0,0.7}

\title{Learning dynamical systems with hit-and-run random feature maps}

\begin{document}
\maketitle
{\normalsize
 \centering
  \authorfootnotes
  Pinak Mandal,\footnote{\thanks{pinak.mandal@sydney.edu.au}} 
  Georg A. Gottwald, \footnote{\thanks{georg.gottwald@sydney.edu.au}}\\
  \textsuperscript{}University of Sydney, NSW 2006, Australia \par
  
 }

\begin{abstract}
We show how random feature maps can be used to forecast dynamical systems with excellent forecasting skill. We consider the tanh activation function and judiciously choose the internal weights in a data-driven manner such that the resulting features explore the nonlinear, non-saturated regions of the activation function. We introduce skip connections and construct a deep variant of random feature maps by combining several units. To mitigate the curse of dimensionality, we introduce localization where we learn local maps, employing conditional independence. Our modified random feature maps provide excellent forecasting skill for both single trajectory forecasts as well as long-time estimates of statistical properties, for a range of chaotic dynamical systems with dimensions up to 512. In contrast to other methods such as reservoir computers which require extensive hyperparameter tuning, we effectively need to tune only a single hyperparameter, and are able to achieve state-of-the-art forecast skill with much smaller networks. 
\end{abstract}

\section{Introduction}\label{sec:intro}

Data-driven modelling of complex dynamical systems has sparked much interest in recent years, with remarkable success in, for example, weather forecasting, producing comparable or even better results than traditional operational equation-based forecasting systems \cite{BiEtAl23,LamEtAl23,PriceEtAl24}. Predicting chaotic dynamical systems with their inherent sensitivity to initial conditions is a formidable challenge. Direct numerical simulation of the underlying dynamical systems often requires small time steps and high spatial resolution due to the presence of multi-scale phenomena; moreover, the underlying equations may not even be known for some complex systems and scientists have to face a certain degree of model error. Substituting costly direct simulation of the underlying dynamical system by a surrogate model which is learned from data is an attractive alternative. Scientists have adopted recurrent networks as their go-to architecture for mimicking dynamical systems. \textcolor{red-}{Remarkably, more complex architectures such as Long Short-Term Memory (LSTM) architectures \cite{HochreiterSchmidhuber97} have been replaced by much simpler architectures such as reservoir computers (RC) or Echo-State Networks (ESN) \cite{MaassEtAl02,Jaeger02,JaegerHaas04}, exhibiting better forecasting capabilities with forecasting times exceeding several Lyapunov units \cite{vlachas2020backpropagation,BompasEtAl20}. Indeed, reservoir computing has emerged as the prominent architecture for modeling and predicting the behavior of chaotic dynamical systems \cite{pathak2018model, rafayelyan2020large,nakajima2021reservoir,LevineStuart22,platt2022systematic}.} Its appeal lies in the ability to process complex, high-dimensional data with relatively simple training procedures. Recently, it was shown that RCs can be further simplified in a variant resembling nonlinear vector autoregression machines, requiring fewer hyperparameters \cite{GauthierEtAl21,Bollt21}. 

We consider here an even simpler version of RCs, which eliminates the internal dynamics of the reservoir and hence requires fewer parameters. These well known random feature maps (RFMs) \cite{RahimiRecht08} can be viewed as a single-layer feedforward network in which the internal weights and biases are fixed, and the outer weights are determined by least-square regression. This approach simplifies the training process and reduces computational costs compared to fully trainable recurrent networks. RFMs have recently been shown to perform very well for learning dynamical systems \cite{GottwaldReich21,NelsenStuart21,LevineStuart22, mandal2024choice}. RFMs enjoy the universal approximation property, and can, in principle, approximate any continuous function arbitrarily well \cite{Cybenko89,ParkSandberg91,Barron93,RahimiRecht08b}. This, however, does not tell a practitioner how to construct a random feature map model so that it well approximates smooth functions, and in particular how to optimally choose the internal weights. Indeed, the performance of RFMs is sensitive to the random but fixed internal weights. Recently there has been interest in finding approximate methods to choose the internal parameters to increase the forecasting capabilities of random feature maps \cite{LevineStuart22,DunbarEtAl24,mandal2024choice}. We follow here our strategy developed in \cite{mandal2024choice} designed for $\tanh$-activation functions, and employ a hit-and-run algorithm to initialize the non-trainable internal parameters ensuring that for the given training data the weights do not project the data into either the saturated region of the $\tanh$-function or the approximately linear region. In the former case, the RFM would not be able to distinguish different data points whereas in the latter case the RFMs would reduce to a linear model which would not be able to capture a nonlinear dynamical system. 

In addition, we introduce several modifications to the classical RFMs. Rather than learning the propagator map we formulate the learning problem to estimate the vector field instead. This is similar to skip connections in residual networks \cite{HeEtAl16} and has recently been used in RCs \cite{CeniGallicchio24}. We then formulate a deep variant of RFMs by constructing a succession of different RFMs that are individually trained. Together with the skip connection, this construction resembles an Euler discretization of a neural ODE \cite{E17}. A similar construction of multi-step learning has been applied to ESNs for forecasting \cite{akiyama2022computational} and classification problems \cite{DingEtAl15,UzairEtAl18}. RFMs suffer, like all kernel methods, from a curse of dimensionality, requiring an exponentially increasing amount of data for increasing dimension to achieve a specified degree of accuracy. To mitigate the curse of dimensionality we employ a localization scheme, assuming that in typical dynamical systems interactions are local and the learning problem can be restricted to a smaller dimensional local region rather than globally for the whole state space. Localization has the additional computational advantage of being parallelizable. Localization schemes have previously been applied to RCs, LSTMs, and generative models \cite{pathak2018model,vlachas2020backpropagation,platt2022systematic,GottwaldReich24}. 

We evaluate our RFMs and the various modifications on three benchmark systems of increasing complexity: the $3$-dimensional Lorenz-63 model, the $40$-dimensional Lorenz-96 model and the Kuramoto-Sivashinsky equation as an example of a partial differential equation. These systems highlight the versatility of random feature models, which achieve state-of-the-art forecasting performance with one or more orders of magnitude fewer parameters and lower computational cost compared to RCs, making them powerful tools for prediction and analysis. We shall see that the width of the RFM needs to be sufficiently large in order to produce reliable features. Once RFMs are of a sufficiently large width, the forecasting performance of RFMs is increased more by increasing depth rather than increasing the width (when the total number of parameters is kept fixed).\\ 

The paper is organized as follows. In Section~\ref{sec:method}, we describe the RFM framework, its deep and local extensions along with the performance metrics used to evaluate our surrogate models. In Section~\ref{sec:results} we show that RFMs are capable of producing accurate forecasts for single trajectories as well as accurate estimates of the long-time statistical properties of the underlying dynamical systems. We provide a comparison with benchmark results from recent literature. Finally, in Section~\ref{sec:conclude}, we close with a brief summary and discussion.
\section{Methodology}
\label{sec:method}

We consider a $D$-dimensional dynamical system which is observed at discrete times $t_n=n\Delta t$ with constant sampling time $\Delta t$. Given $N+1$ observations $\bb{u}_0, \bb{u}_1, \bb{u}_2,\cdots, \bb{u}_{N}$ with $\bb{u}_n=\bb{u}(t_n)$, our goal is to construct a surrogate model ${\bb \Psi}_{\Delta t}$ that approximates the map ${\bb{\Psi}}: \bb{u}_n\mapsto \bb{u}_{n+1}$ of the underlying dynamical system as closely as possible. We assume that our observations are complete and noise-free. We employ here random feature maps to construct the surrogate models. We begin with a description of classical random feature maps before introducing our modifications, namely skip connections and deep and localized variants.

\textcolor{orange-}{We remark that the framework of RFMs can be extended to deal with the more realistic scenario of partial and noisy observations. Observational noise, which when untreated has a detrimental effect on learning with RFMs, can be controlled by combining the RFM learning task with an ensemble Kalman filter \cite{gottwald2021supervised}. To overcome the implied non-Markovianity of a partially observed dynamical system, time-delay embedding techniques can be employed \cite{GottwaldReich21}. We do not consider these extensions here and restrict to noise-free and complete observations which allows for better  benchmarking.}


\subsection{Classical random feature maps }
\label{ssec:rfm}

Random feature maps are feedforward neural networks consisting of an internal layer of width $D_r$ and an external layer. \textcolor{orange-}{We use $\tanh$ as the activation function for the internal layer.} The weights $\bb{W}_{\rm in}$ and biases $\bb{b}_{\rm in}$ of the internal layer are drawn from some user-defined distribution and are kept fixed. The external layer weights $\bb{W}$ are learned. An RFM is compactly written as
\begin{align}
\bb{u} \mapsto \bb{W}\tanh{(\mathbf{W}_{\rm in}\mathbf{u} + \mathbf{b_{\rm in}})},
\label{eq:rfm}
\end{align}
where $\bb{u}\in\mathbb{R}^D, \bb{W}_{\rm in}\in\mathbb{R}^{D_r\times D}$, $\bb{b}_{\rm in}\in\mathbb{R}^{D_r}$, and $\bb{W}\in\mathbb{R}^{D\times D_r}$. The surrogate map 
\begin{align}
{\bb\Psi}_{\Delta t}(\bb{u}_{n}) = \bb{W}\tanh{(\mathbf{W}_{\rm in}\mathbf{u}_{n} + \mathbf{b_{\rm in}})}
\end{align}
provides an estimate for the observed $\bb{u}_{n+1}$. Training RFMs amounts to training the external layer by minimizing the following regularized cost,
\begin{align}
     \underset{\mathbf{W}}{\argmin}\;\|\mathbf{W}{\mathbf{\Phi}}(\bb{U})-\bb{U}^\prime\|_F^2 + \beta\|\mathbf{W}\|_F^2
\label{eq:cost},
\end{align}
where $\bb{U}\in\mathbb{R}^{D\times N}$ contains the observations $\{\bb{u}_n\}_{n=0}^{N-1}$ across its columns and $\bb{U}^\prime \in\mathbb{R}^{D\times N}$ contains the time-shifted observations $\{\bb{u}_{n+1}\}_{n=0}^{N-1}$ across its columns. The feature matrix ${\mathbf{\Phi}}(\bb{U})$ denotes the output of the internal layer computed as $\bb{u}\mapsto\tanh(\mathbf{W}_{\rm in}\mathbf{u} + \mathbf{b_{\rm in}})$. The regularization parameter $\beta$ is a hyperparameter which requires tuning. Here $\|\cdot\|_F$ denotes the Frobenius norm. The solution to the ridge regression problem \eqref{eq:cost} is explicitly given by 
\begin{align}
\mathbf{W} = \bb{U}^\prime {\mathbf{\Phi}}^\top \left( {\mathbf{\Phi}}{\mathbf{\Phi}}^\top + \beta \mathbf{I}\right) ^{-1},
\label{eq:lsq}
\end{align}
where we have omitted the dependency of the feature matrix ${\mathbf{\Phi}}$ on the data $\bb{U}$; in particular, no costly backpropagation is required. The quality of the learned surrogate model sensitively depends on the random initialization of the internal layer and the hyperparameter $\beta$.  


\subsection{Initialization of the internal layer}
\label{ssec:sample} 

\begin{figure}[!htp]
    \centering
    \includegraphics[width=0.6\linewidth]{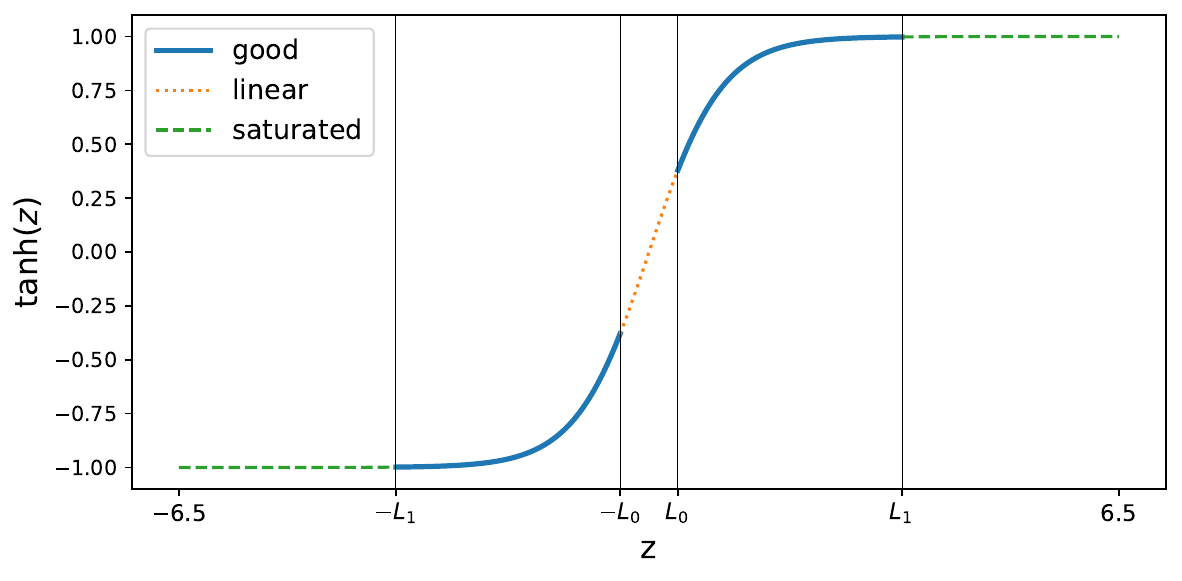}
    \caption{Illustration of the types of features produced by a $\tanh$-activation function, motivating the choice of the internal weights and biases $(\mathbf{W}_{\rm in},\mathbf{b_{\rm in}})$. Here and elsewhere $L_0=0.4$ and $L_1=3.5$.}
    \label{fig:tanh}
\end{figure}

We briefly describe the effective sampling scheme for the internal layer introduced in our previous work \cite{mandal2024choice}, which is used throughout this work. Our algorithm is based on the specific functional form of the $\tanh$-activation function. Consider a row of the internal weight matrix $\mathbf{W}_{\rm in}$ which we denote by $\mathbf{w}_{\rm in}\in \mathbf{R}^{D}$, and an entry of the bias vector which we denote by $b_{\rm in}$. The domain of the $\tanh$-activation function has three distinct regions: a saturated region, a linear region and the complement of these two, as illustrated in Figure~\ref{fig:tanh}. Internal weights for which the features $\phi(\bb{u})=\tanh({\mathbf{w}}_{\rm in}\bb{u} + b_{\rm in})$ are saturated i.e. $\phi(\bb{u})\approx \pm 1$ or equivalently $|{\mathbf{w}}_{\rm in}\bb{u} + b_{\rm in}| \ge L_1$ (we use $L_1=3.5$ throughout) are clearly bad choices as the RFMs would not be able to distinguish between different input signals. Internal weights for which the features lie in the linear region with $|{\mathbf{w}}_{\rm in}\bb{u} + b_{\rm in}| \le L_0$ (we use $L_0=0.4$ throughout), lead to a linear model, which is undesirable for learning nonlinear systems. We hence aim to draw internal weights such that the associated features are neither saturated nor linear for any of the training data, these features are labelled as \textit{good} in Figure~\ref{fig:tanh}. The method proposed in \cite{mandal2024choice} achieves this by a hit-and-run algorithm: starting from a feasible solution $\mathbf{w}_{\rm in}=0$ and $b_{\rm in}$ uniformly sampled from the interval $ \pm[L_0,L_1]$ we pick random directions in a convex set determined by the training data and the inequalities $L_0<{\mathbf{w}}_{\rm in}\bb{u} + b_{\rm in} < L_1$ or $-L_1<{\mathbf{w}}_{\rm in}\bb{u} + b_{\rm in} < -L_0$. Determining where the line segment defined by this direction intersects the convex set allows us to sample weights that map the training data to the aforementioned good features.  This process is repeated until $D_r$ independent rows $\mathbf{w}_{\rm in}$ and biases ${b}_{\rm in}$ are drawn. We stress that the hit-and-run algorithm does not perform any training by optimization but simply samples the internal weights from a data-informed convex set.

The method is summarized in Appendix~\ref{sec:app:algo:hr} in Algorithm~\ref{algo:hr}; for a detailed discussion regarding the geometry of the algorithm we refer to \cite{mandal2024choice}. We emphasize that $L_0$ and $L_1$ are treated as constants in our approach. While the selection of their values to delineate good features from bad features could, in principle, be considered hyperparameters requiring tuning, we observe no significant changes in the forecasting capabilities of the learned surrogate maps for values close to $L_0 = 0.4$ and $L_1 = 3.5$. Consequently, we have consistently used $L_0 = 0.4$ and $L_1 = 3.5$ for all the experiments presented in this work.

\subsection{Skip connections}
\label{ssec:skip} 

A simple but effective modification of the random feature map is the introduction of a skip connection from the input to the output \cite{HeEtAl16,CeniGallicchio24}. \textcolor{blue-}{In particular, we learn the tendency map $\bb{F}_{\Delta t}: \bb{u}_{n}\mapsto\bb{u}_{n+1}-\bb{u}_{n}$ with an RFM, rather than the propagator map ${\bb \Psi}_{\Delta t}: \bb{u}_{n}\mapsto\bb{u}_{n+1}$. We hence solve the least-square problem \eqref{eq:cost} where now $\bb{U}^\prime \in\mathbb{R}^{D\times N}$ contains the observed tendencies $\{\bb{u}_{n+1} - {\bb{u}}_n \}_{n=0}^{N-1}$ across its columns, with solutions given by \eqref{eq:lsq}. We will refer to this variant of RFM as SkipRFM.}

\textcolor{blue-}{Bar the constant factor of $\Delta t$, SkipRFM can be viewed as learning a single Euler step in a forward-Euler discretization 
\begin{align}
\bb{u}_{n+1} = \bb{u}_{n} +\bb{F}_{\Delta t}(\bb{u}_n) = \bb{u}_{n} + \Delta t \;\bb{\tilde F}_{\Delta t}(\bb{u}_n) ,
\end{align}
for a dynamical system with the vector field 
\begin{align}
\bb{\tilde F}_{\Delta t}(\bb{u}_n) = \frac{1}{\Delta t} \bb{W}\tanh{(\mathbf{W}_{\rm in}\mathbf{u}_{n} + \mathbf{b_{\rm in}})}.
\end{align}
Note that the map $\bb{F}_{\Delta t}$ is learned from data with a fixed and specified value of the sampling time $\Delta t$. When applying the learned map to forecast unseen data, we show results for the same value of $\Delta t$ we used for learning. Choosing different values, which would be possible for actual numerical integrators, can lead to instabilities \cite{KrishnapriyanEtAl23,LevineStuart22}, significantly deteriorating the forecast capabilities of the learned model. We, however, empirically found our surrogate models to work well when trained on data of finer temporal resolution compared to the testing data for moderate ratios of sampling times of training and testing data.}

RFMs with skip connections tend to be marginally better at forecasting than those without skip connections. In fact, in all our test cases, models with skip connections achieved the highest forecast times, as we will see in Section~\ref{sec:results}.

%


\subsection{Deep random feature maps}
\label{ssec:deep} 

\begin{figure}
    \centering
    \includegraphics[scale=0.58]{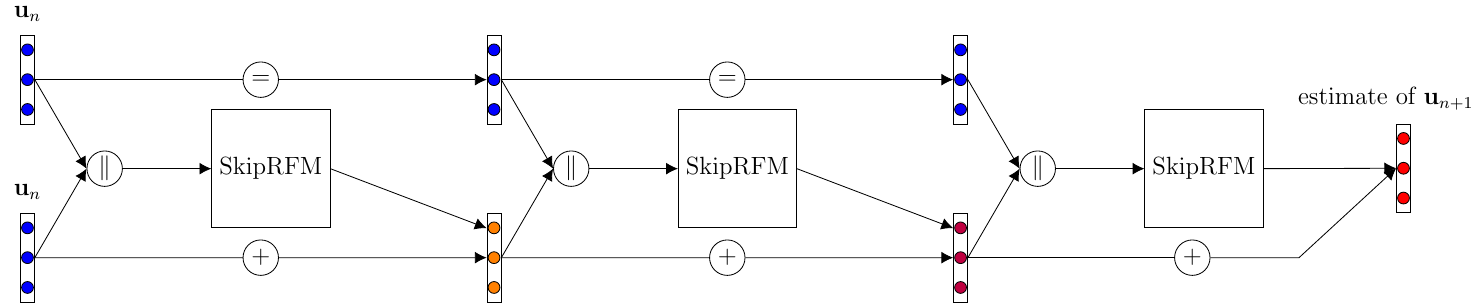}
    \caption{Schematic of the deep architecture DeepSkip with depth $B=3$. The symbols $\|$, $=$ and $+$ denote concatenation, identity operation and addition (skip connection), respectively.}
    \label{fig:DeepSkip}
\end{figure}

We now increase the complexity of random feature models by chaining multiple units together to construct deep models and explore some of their benefits. Figure~\ref{fig:DeepSkip} provides an outline of a deep model. \textcolor{blue-}{We initialize the input with two copies of the state $\bb{u}_n$ at time $t_n$, which are concatenated, to form
\begin{align}
{\bb{y}}^{(0)}_n = 
\begin{bmatrix} 
\bb{u}_n\\ 
\bb{u}_n
\end{bmatrix}.
\end{align}
This augmented state is passed through the first single random feature model unit. The output of the first single unit (and of all following units) replaces one half of the augmented state to form
\begin{align}
{\bb{y}}^{(\ell)}_n = 
\begin{bmatrix} 
\mathbf{W}^{(\ell)}\tanh( \mathbf{W}_{\rm in}^{(\ell)} \bb{y}^{(\ell-1)}_n + \mathbf{b}_{\rm in}^{(\ell)} ) \\ 
\bb{u}_n
\end{bmatrix},
\label{eq:augy}
\end{align}
and the updated augmented state is again passed through the next unit and so on. Here $\mathbf{W}_{\rm in}^{(\ell)}\in \mathbb{R}^{D_r\times2D}$ and $\mathbf{b}_{\rm in}^{(\ell)} \in \mathbb{R}^{D_r}$ with $\ell=1,\ldots,B$ denote the inner weights and biases of the $\ell^{\text{\tiny{\rm{th}}}}$ unit. 
Similarly, $\mathbf{W}^{(\ell)}\in\mathbb{R}^{D\times D_r}$ denotes the outer weight of the $\ell^{\text{\tiny{\rm{th}}}}$ unit which are learned sequentially by solving the least-square problem  
\begin{align}
\underset{\mathbf{W}^{(\ell)}}{\argmin}\;\|\mathbf{W}^{(\ell)}{\mathbf{\Phi}}(\bb{Y}^{(\ell)})-\bb{U}^\prime\|_F^2 + \beta\|\mathbf{W}^{(\ell)}\|_F^2 , 
\label{eq:costB}
\end{align}
where $\bb{Y}^{(\ell)}\in\mathbb{R}^{2D\times N}$ contains $\{\bb{y}^{(\ell)}_n\}_{n=0}^{N-1}$ across its columns. We consider the case when each unit is a standard RFM with $\bb{U}^\prime \in\mathbb{R}^{D\times N}$ containing the time-shifted observations $\{\bb{u}_{n+1}\}_{n=0}^{N-1}$ across its columns, as well as the case when each unit is a SkipRFM unit with $\bb{U}^\prime \in\mathbb{R}^{D\times N}$ containing the tendencies $\{ \bb{u}_{n+1} - \bb{u}_{n}\}_{n=0}^{N-1}$ across its columns. This process is repeated until we go through the final unit with $\ell=B$ and the final updated upper half of the augmented state is our approximation of the state $\bb{u}_{n+1}$ (or  $\bb{u}_{n+1} -  \bb{u}_{n}$ when SkipRFMs are considered) at time $t_{n+1}$.  
When the unit is an RFM, the resulting deep model is referred to as DeepRFM. Similarly, when the unit is a SkipRFM, the corresponding deep model is referred to as DeepSkip. 
We found empirically that using augmented states \eqref{eq:augy}, such that each unit has the state $\bb{u}_n$ as part of its input, rather than using ${\bb{y}}^{(\ell)}_n = \mathbf{W}^{(\ell)}\tanh( \mathbf{W}_{\rm in}^{(\ell)} \bb{y}^{(\ell-1)}_n + \mathbf{b}_{\rm in}^{(\ell)} )$ with $\bb{y}^{(0)}_n=\bb{u}_n$, leads to better performing surrogate models. We also found that solving a regression problem at each unit rather than a single regression problem at the last unit i.e. ${\bb{y}}^{(\ell)}_n = \tanh( \mathbf{W}_{\rm in}^{(\ell)} \bb{y}^{(\ell-1)}_n + \mathbf{b}_{\rm in}^{(\ell)} ),\; (l=1,2,\cdots, B-1)$ with $\bb{y}^{(0)}_n=\bb{u}_n$ and $\bb{y}_n^{(B)}=\bb{W}\bb{y}_n^{(B-1)}$, produces better models. The non-trainable internal weights $\mathbf{W}_{\rm in}^{(\ell)}$ and $\mathbf{b}_{\rm in}^{(\ell)}$ are determined for all units with the hit-and-run Algorithm~\ref{algo:hr} using the same input data $\bb{Y}^{(0)}$. It suffices to tune the regularization hyperparameter $\beta$ in deep RFM architectures for a single unit and reuse it for all the units.}

Similar constructions have recently been used in the context of echo state networks \cite{DingEtAl15,UzairEtAl18,akiyama2022computational}, \textcolor{blue-}{and universal approximation theorems for a different version of a deep RFM were recently proved \cite{BoschEtAl23}.} \textcolor{blue-}{Our deep random feature architecture updates the outer weights sequentially based on the errors incurred at the prior units and is hence reminiscent of stacked boosting \cite{ShapireFreund,KimMyoung03} in machine learning.}

Deep versions of random feature models exhibit improved forecasting capabilities when compared to their shallow counterparts, as will be shown in Section~\ref{sec:results}. Moreover, depth has significant computational advantages. A major benefit of introducing depth is that it allows us to train larger models on a GPU with fixed memory. The total number of weights and biases in a model, henceforth referred to as the model size $S$, significantly influences the model's forecasting skill. But the total memory occupied on the GPU during training of deep RFM models primarily depends on the model width $D_r$ and the size of training data $N$, and not on the model size. This is because we train the constituent units sequentially and hence the GPU needs to handle only one linear regression problem at a time. Therefore, a shallow and a deep model with the same width roughly occupy the same amount of GPU memory during training despite the deeper model having a larger size. Furthermore, introducing depth allows for a significant speed up of training. For a shallow and a deep model of the same size, the deep model necessarily has a smaller width. Therefore, when trained on the same amount of data, the deeper model requires solving regression problems of smaller size. Consequently, among models of the same size, deeper models can train up to an order of magnitude faster, as we will see in Section~\ref{ssec:time}. 
 
We remark that the frequency of observations, or the temporal resolution of the data $\Delta t$, plays a crucial role in determining the forecast skill of a trained surrogate model. Generally, smaller values of $\Delta t$ enable better learning of the underlying dynamical system. In Section~\ref{ssec:L63}, we present an example where shallow models struggle with large $\Delta t$, while deep models demonstrate superior performance.

\subsection{Localization} 
\label{sssec:scheme}

\begin{figure}
    \centering
    \includegraphics[scale=0.8]{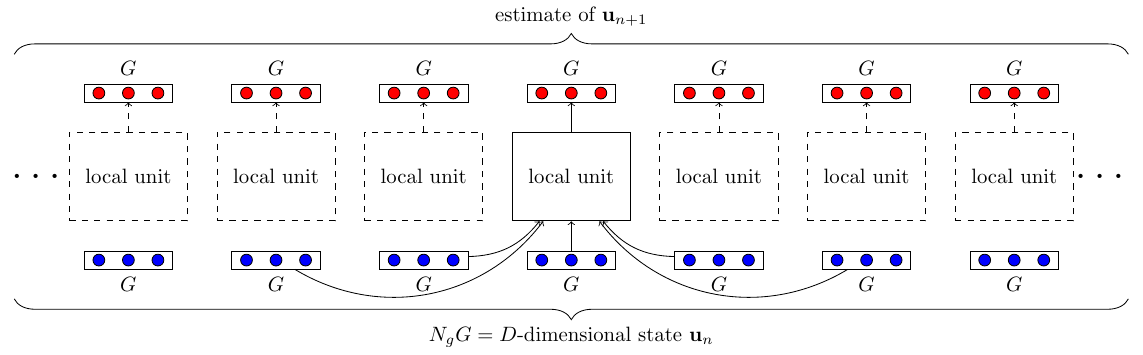}
    \caption{Schematic of a localized architecture. In this example, the local state dimension is $G=3$ and the interaction length is $I=2$.}
    \label{fig:local}
\end{figure}

To mitigate the curse of dimensionality associated with high-dimensional systems with large $D$, we design localized variants of random feature models. Typically in high-dimensional systems, for sufficiently small sampling times $\Delta t$, the state of a variable at future time $t_{n+1}$ does not depend on all other variables at the current time $t_{n}$. An example comes from weather forecasting where the weather at one location typically does not depend on the weather at locations which are several thousand kilometres away. Localization techniques have been successfully employed recently for RCs and LSTMs \cite{pathak2018model, vlachas2020backpropagation, platt2022systematic}. Here we set out to learn $N_g$ localized models by subdividing the state vector into $N_g =  D/G$ local states of dimension $G$ each. For each local vector of dimension $G$ we train a local random feature unit. Each local unit takes its own local state along with the states of its neighbours as input, aiming to predict its own local state at the next time step. Concretely, we assume that the state of a local region at time $t_{n+1}$ depends on the state of the same local region as well as on its $2I$ neighbouring local regions at time $t_n$, where $I$ is called the interaction length. The pair $(G, I)$ defines a localization scheme. Figure~\ref{fig:local} illustrates the structure of a localized random feature model. For shallow localized models the input dimension for each unit is $(2I+1)G$. For deep localized models, instead of doubling the input dimension, we augment the input of a local unit with only its own local state giving us an input dimension of $2(I+1)G$. Localized variants of RFM, SkipRFM, DeepRFM and DeepSkip are coined LocalRFM, LocalSkipRFM, LocalDeepRFM and LocalDeepSkip, respectively. We indicate the localization scheme in the subscript e.g. a LocalDeepSkip model utilizing local state dimension $G=4$ and interaction length $I=2$ is referred to as LocalDeepSkip$_{4, 2}$. A good localization scheme is crucial for the success of a localized model. Appendix~\ref{ssec:loc} explores various localization schemes for our test problems and provides some general guidelines for selecting an optimal localization scheme. 

Besides controlling the curse of dimensionality, localization also allows for a considerable computational advantage via a tensorized implementation. For dynamical systems with translational symmetry such as the Lorenz-96 system and the Kuramoto-Sivashinsky equation which we consider in Section~\ref{sec:results}, all local units within the complete architecture can be chosen to be identical, allowing us to train a single unit and replicate the trained parameters across the entire model. We exploit this a step further by working with only a single unit that processes information from the entire state using matrix-tensor operations, producing the complete state vector for the next time step. This approach eliminates the need to store multiple local units, reducing the model size by a factor of $N_g$. The reduction in input dimension allows us to accommodate localized models with much larger width $D_r$ compared to their non-localized counterparts. This, in turn, allows localized models to be significantly more expressive. In Section~\ref{sec:results}, we see that the localized models far outperform the non-localized models in the high-dimensional test cases.


\subsection{Dealing with possibly ill-conditioned data}
\label{ssec:noise}

The data $\bb{U}$ may be ill-conditioned, for example, subsequent snapshots of a partial-differential equation may only vary significantly in a small region for a sufficiently small sampling time $\Delta t$. For simplicity, let us assume that we are employing an RFM to learn a dynamical system. The outer weight matrix $\bb{W}$ depends on the training data $\bb{U}$, and as a result, is also ill-conditioned. If the condition number of $\bb{W}$ is too large, then the learned surrogate model becomes unstable if run in autonomous mode for the test data. Indeed, during multiple recursive applications of the surrogate model small errors accumulate leading to the predicted state departing from the attractor. The internal parameters $(\bb{W}_{\rm in},\bb{b}_{\rm in})$, sampled by Algorithm~\ref{algo:hr} (see Appendix~\ref{sec:app:algo:hr}), are then unable to produce good features, and further recursions typically lead to numerical blow-up. 

We mitigate such instabilities by artificially adding small noise to the training data. Indeed, adding small noise to an ill-conditioned matrix has been shown rigorously to produce a well-conditioned matrix with high probability \cite{spielman2004smoothed}. The added artificial noise on the data matrix $\bb{U}$ reduces the condition number of the training data and, consequently, that of the outer weight matrix $\bb{W}$. The noise should be sufficiently small as not to contaminate the signal and ensure no degradation of the accuracy of the one-step surrogate map. We found that noise, for which the noisy and the original noise-free data are indistinguishable by eye, is sufficient to control instability while still providing accuracy of the learned surrogate model. This strategy is relevant to all the architectures covered in this section. In Section~\ref{ssec:KS} we show an example of ill-conditioned training data and its catastrophic effect on the forecast skill of a LocalDeepSkip model. In the following, we distinguish the models trained on data with added artificial noise by appending 'N' to their name, e.g. a LocalDeepSkip model trained on noisy data is referred to as LocalDeepSkipN. \textcolor{orange-}{Adding noise to training data has been used routinely and unconditionally for learning deterministic dynamical systems with LSTMs and RCs \cite{VlachasEtAl18,vlachas2020backpropagation}. We only apply noise when dealing with ill-conditioned training data $\bb{U}$.}


\subsection{Performance metrics}
\label{ssec:metric}

To evaluate the forecast skill of our surrogate models we test them on unseen test data. We initialize the surrogate model with the initial condition of a noise-free test trajectory, and then let the model run in autonomous mode according to 
\begin{align}
\label{eq:auton}
{\hat{\bb{u}}}_{n+1}={\bb \Psi}_{\Delta t}({\hat{\bb{u}}}_n) 
\end{align}
with $\hat{\bb{u}}_0 = \bb{u}_0$. Note that here $\bb{u}$ denotes test data. For simplicity, we label test data the same way as training data when there is no danger for confusion. We compare the surrogate forecasts ${\hat{\bb{u}}}_n$ with the test data $\bb{u}_n$. To quantify the forecast skill we compute the valid prediction time (VPT), measured in Lyapunov times, 
\begin{align}
    {\rm VPT} = \frac{1}{T_{\Lambda}} \underset{n}{\sup} \left\{ n \Delta t : \sqrt{\frac{1}{D}\sum_{i=1}^D\left( \frac{\hat{\bb{u}}_{n,i} - \bb{u}_{n,i}}{\sigma_i} \right)^2} < \varepsilon \right\},
\end{align}
where $T_{\Lambda}=1/\Lambda$ is the Lyapunov time with $\Lambda$ being the maximal Lyapunov exponent. The data mismatch is normalized componentwise by the standard deviation $\sigma\in\mathbb{R}^D$. The standard deviation is numerically estimated from the training data.  The parameter $\varepsilon>0$ is a chosen error threshold. VPT is a diagnostic which has been used for RCs and LSTMs and allows us to compare with several benchmark results from the literature. To obtain meaningful results with a statistical significance we run many realizations where we randomly draw the training data, test data and the internal weights. 

We further test the long-term behaviour of the surrogate models by running long simulations and comparing their empirical invariant measures with those of the original dynamical system. To quantify the quality of the long-time statistical behaviour we estimate the Wasserstein distance $W_2$ between the $1$-dimensional empirical marginal distributions under comparison. We have further estimated the power spectral density of the mean state evolution, another popular probe for long-term statistics. However, we found that the power spectral density is too well recovered by all our RFM variants and hence is not suitable to study their relative performance. We therefore only report on the empirical histograms.


\subsection{Data and code}
\label{ssec:code}

The code for reproducing the results shown here and the forecast data are openly available on Github at \url{https://github.com/pinakm9/DeepRFM}. The code is written in Python and heavily utilizes PyTorch for implementation of the random feature models as well as a parallelized version of Algorithm~\ref{algo:hr} (see Appendix~\ref{sec:app:algo:hr}).
\section{Results}
\label{sec:results}

We evaluate our random feature surrogate models on three widely-used benchmark dynamical systems: the $3$-dimensional Lorenz-63 system, the $40$-dimensional Lorenz-96 system and the Kuramoto-Sivashinsky equation as an example of a partial differential equation which we discretize with $512$ gridpoints. For all three systems we ensure that the training data and the test data evolve on the attractor by running simulations of the original dynamical system for a sufficiently long time. 

To obtain meaningful statistics of the forecast performance metric VPT we generate $500$ random realizations differing in the training data, the testing data and the non-trainable internal weights and biases of the surrogate model. For each model, the regularization hyperparameter $\beta$ was optimized via grid search. 

To compare empirical histograms obtained from long-time simulations, Wasserstein distances $W_2$ are estimated from $3\times10^4$ random samples for each model using the Sinkhorn algorithm with an entropy regularization parameter of $10^{-2}$ \cite{feydy2019interpolating}. 

All experiments were done on the A100 GPU provided by Google Colab. Additional numerical details regarding the results shown here can be found in Appendix~\ref{ssec:detail}. 


\subsection{Lorenz-63}
\label{ssec:L63} 
 
In this section we demonstrate the forecast skill and long-term behavior of surrogate models for the Lorenz-63 (L63) system with standard parameters \cite{lorenz1963deterministic}, 
\begin{align}
    \begin{aligned}
    \frac{dx}{dt} &= 10 (y - x), \\
    \frac{dy}{dt} &= x (28 - z) - y, \\
    \frac{dz}{dt} &= x y - \frac{8}{3} z.
    \end{aligned}\label{eq:L63}
\end{align} 
The maximal Lyapunov exponent is estimated to be $\Lambda = 0.91$ \cite{platt2022systematic}. Localization is not required for this low-dimensional system, and we consider here the non-localized versions RFM, SkipRFM and DeepSkip. 
 
\begin{figure}[!htp]
    \centering
    \includegraphics[scale=0.5]{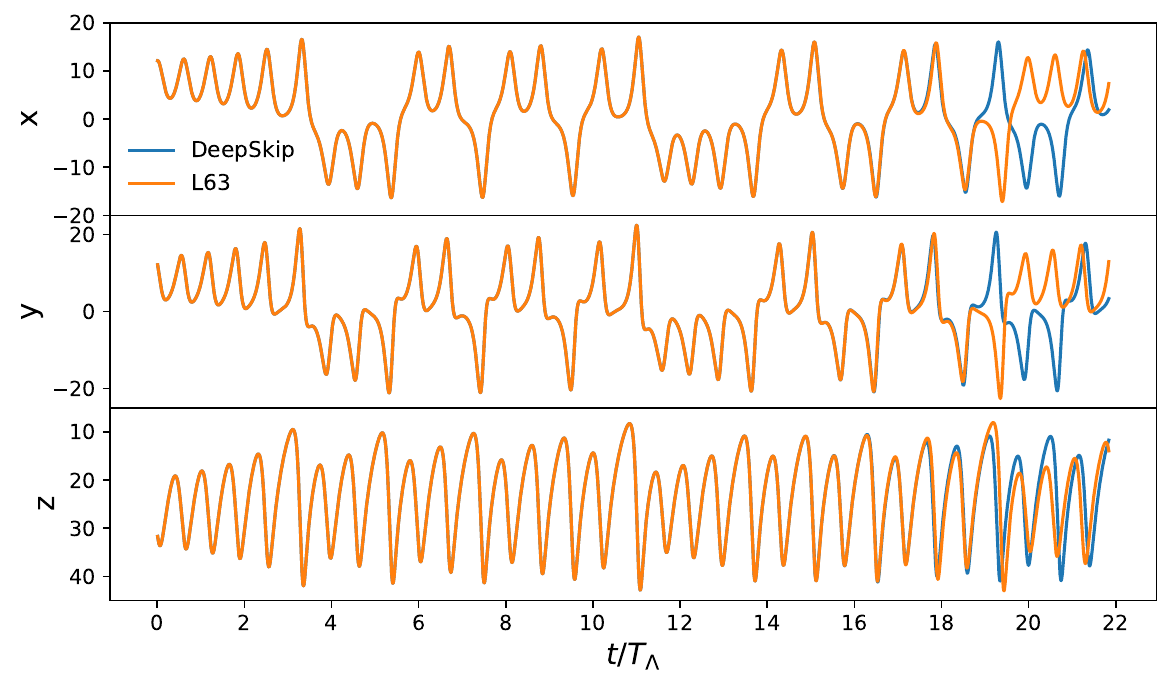}
    \caption{An example of a forecast by a DeepSkip model with width $D_r=1,024$ and depth $B=16$ for the L63 system \eqref{eq:L63}. The surrogate model is able to forecast accurately up to ${\rm{VPT}}\approx 19$ Lyapunov time units.}
    \label{fig:L63-forecast}
\end{figure}

\begin{figure}[!htp]
    \centering    \includegraphics[width=\linewidth]{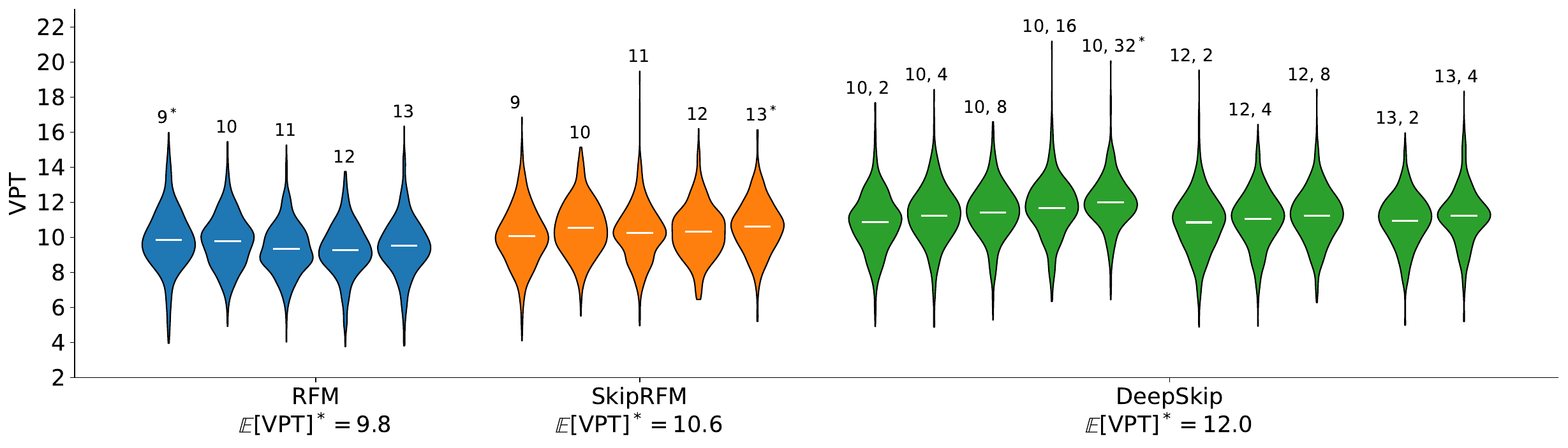}
    \caption{Kernel density plots of VPT for the L63 system \eqref{eq:L63} for $(N, \Delta t, \varepsilon)=(5\times10^4, 0.01, 0.3)$. For RFM and SkipRFM, $\log_2(D_r)$ is indicated on the top of the plots. For DeepSkip, $(\log_2(D_r), B)$ is indicated on the top of each plot. The $*$-symbol indicates the model with the best mean VPT within each architecture.}
    \label{fig:L63-1}
\end{figure}

\begin{figure}[!htp]
    \centering
    \includegraphics[scale=0.5]{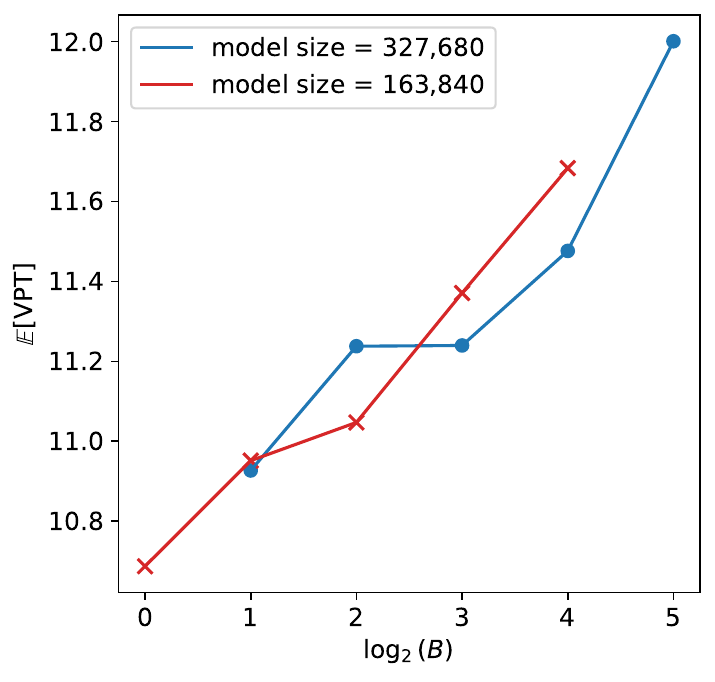}
    \caption{Mean VPT for DeepSkip as a function of depth $B$ for constant model size $S$ for the L63 system \eqref{eq:L63}. Along each curve the model size  $S=(3D+1)D_rB$ remains constant and the width $D_r$ decreases with depth.}
    \label{fig:L63-depth}
\end{figure}

Figure~\ref{fig:L63-forecast} shows a sample forecast of a DeepSkip model which is accurate up to approximately ${\rm{VPT}}\approx19$ Lyapunov time units. However, there is a significant variability in the VPT due to the sensitivity to initial conditions of the chaotic L63 system. Figure~\ref{fig:L63-1} shows the distributions of VPT for training data of length $N=5\times10^4$ sampled with $\Delta t=0.01$ and a VPT error threshold value of $\varepsilon=0.3$. We show results for RFM, SkipRFM and DeepSkip. It is seen that increasing the width $D_r$ past $D_r = 512$ does not lead to an improvement of the mean forecast VPT for the shallow versions RFM and SkipRFM, which saturate around $\mathbb{E}[{\rm{VPT}}]\approx 9.6$ for RFM and slightly higher with  $\mathbb{E}[{\rm{VPT}}]\approx 10.5$ for SkipRFM. On the other hand, increasing the depth $B$ consistently improves the performance of DeepSkip for each fixed width $D_r$. The best performing deep models are able to forecast approximately $1.4$ Lyapunov time units longer compared to the best performing shallow models. The best mean forecast VPT is achieved for $D_r=1,024$ and depth $B=32$ with $\mathbb{E}[{\rm{VPT}}]= 12$. Deep models improve with depth even when the model size $S=(3D+1)D_rB$ is kept fixed, as seen in Figure~\ref{fig:L63-depth} for two different model sizes. Since depth allows us to train larger models as discussed in Section~\ref{ssec:deep}, we are able to train deep models that are $3$ times larger than the largest shallow model increasing the expressivity of the model. In Appendix~\ref{ssec:time} we show that deep architectures allow for an order of magnitude faster training. In Appendix~\ref{ssec:detail} a comparison of our variants of the random feature map, \textcolor{blue-}{including DeepRFM}, is shown in Tables~\ref{tab:L63_1_s} and \ref{tab:L63_0_s} for different model sizes, reporting on the mean, median, standard deviation of the VPT as well as the maximum and minimum values. \textcolor{blue-}{DeepSkip performs better than DeepRFM with a $2$ Lyapunov units larger average VPT for $\Delta t=0.01$.}

Table~\ref{tab:L63-lit} shows a comparison of our best performing DeepSkip models with recent benchmark results, highlighting that DeepSkip is able to achieve state-of-the-art forecast times with an order of magnitude smaller model size. 

\begin{table}[!htp]
    \centering
    \begin{tabular}{|c|c|c|c|c|c|c|}
         \hline
         Source & Model & $\log_{10}(\text{model size})$ & $\mathbb{E}[{\rm VPT}]$ & $N$ & $\Delta t$ & $\varepsilon$\\  \hline
         Akiyama et al. (2022)~\cite{akiyama2022computational} & Multi-step ESN & $5.05$ & $9.3$ & $2\times10^4$ & $0.02$ &  $0.4$\\\hline
          Platt et al. (2022)~\cite{platt2022systematic} & RC & $6.60$ & \cellcolor{pink}$11.8-12.0$ & $5\times10^4$ & $0.01$ & $0.3$\\ \hline
         Koster et al. (2023)~\cite{koster2023data} & DI-RC (SINDy) & $6.00$ & $4.0$ & $10^4$ & $0.01$ & $\sqrt{0.4}$\\ \hline
          Our work & DeepSkip RFM & \cellcolor{pink}$5.52$ & \cellcolor{pink}$12.0$ & $5\times10^4$ & $0.01$ & $0.3$\\ \hline
          Our work & DeepSkip RFM & $5.52$ & $11.8$ & $2\times10^4$ & $0.02$ & $\sqrt{0.05}$\\ \hline
    \end{tabular}
    \caption{Comparison of mean VPT and corresponding model sizes from recent benchmark results for forecasting the L63 system \eqref{eq:L63}. The result corresponding to the best mean VPT $\mathbb{E}[{\rm VPT}]$ is reported for each source. The largest $\mathbb{E}[{\rm VPT}]$ and the corresponding smallest model size are highlighted by red shading.}
    \label{tab:L63-lit}
\end{table}

\begin{figure}[!htp]
    \centering
    \includegraphics[scale=0.5]{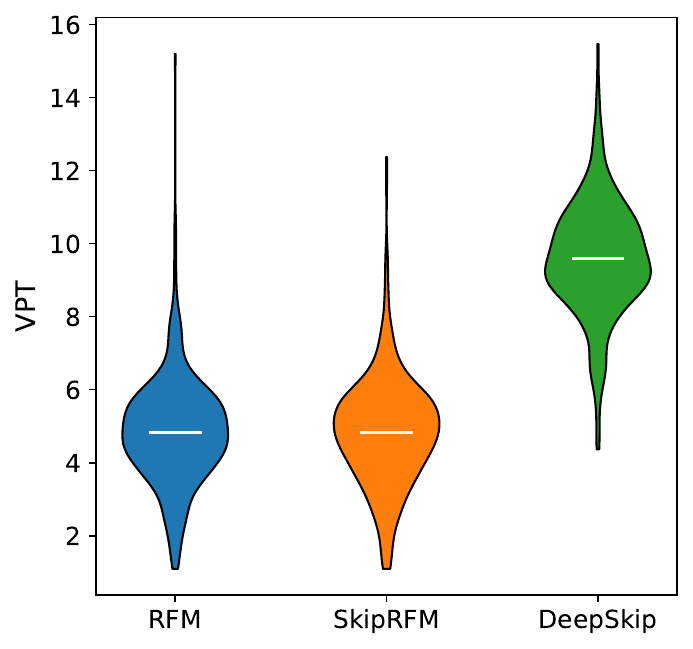}
    \caption{Kernel density plots of VPT for the L63 system \eqref{eq:L63}  for $(N, \Delta t, \varepsilon)=(5\times10^4, 0.1, 0.3)$. Sizes of the models are $S=114,688$, $S=114,688$ and $S=114,560$, from left to right, with,  $D_r=16,384$, $D_r=16,384$ and $D_r=716$ respectively. The DeepSkip model has depth $B=16$.}
    \label{fig:L63-large-dt}
\end{figure}

In general, finer temporal resolution is beneficial for learning the dynamics. In Figure~\ref{fig:L63-large-dt} we see the effect of increasing the sampling time to a fairly large value of $\Delta t = 0.1$, which is about a tenth of a Lyapunov time, on the forecast skill for various models of nearly similar size. The deep model outperforms the shallow models by $\sim$ $4.8$ Lyapunov units on average. The mean VPT drops for RFM from $9.5$ to $4.8$, for SkipRFM from $10.4$ to $4.8$ and for DeepSkip from $11.4$ to $9.6$ when $\Delta t$ is changed from $0.01$ to $0.1$. Smaller sampling times $\Delta t$ allow for a better approximation of temporal derivatives and therefore SkipRFM outperforms RFM for small $\Delta t$. This advantage, however, vanishes at higher $\Delta t$ and both perform equally. For RFM, SkipRFM, and DeepSkip, the mean VPT drops by $49.5\%$, $53.8\%$ and $15.8\%$, respectively, indicating that the deep architectures are the least susceptible to the temporal resolution of the training data.

\begin{figure}[!htp]
    \centering
    \includegraphics[scale=0.5]{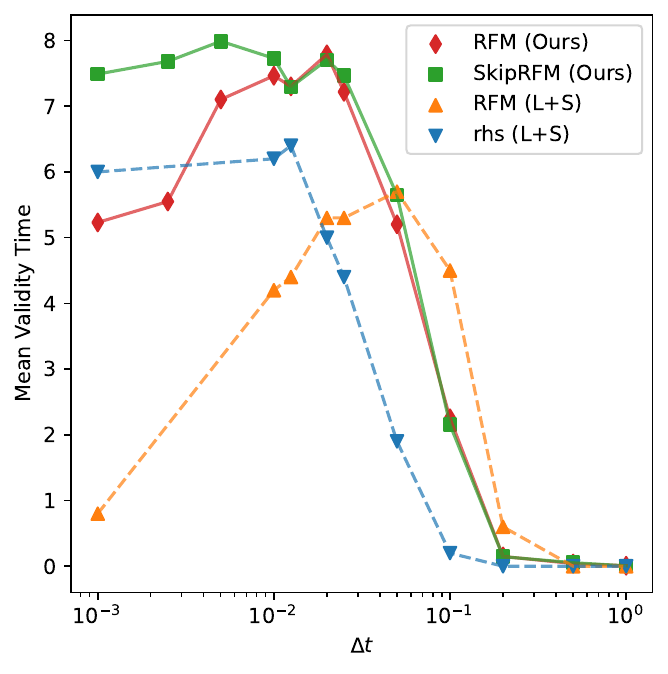}
    \caption{\textcolor{blue-}{Mean validity time $\tau_f$ as a function of the sampling time $\Delta t$ for the L63 system \eqref{eq:L63} with $D_r=200$. Each model was trained on a time-series spanning $T=1,000$ time units and hence the length of the training data $N$ decreases with increasing $\Delta t=T/N$. Averages for our models were computed using $500$ realizations differing in the training data, the testing data and the non-trainable internal weights. The mean validity times labelled as (L+S) are taken from Figure 5 in \cite{LevineStuart22}. Note that we show results for two extra values of $\Delta t=2.5\times10^{-3},5\times10^{-3}$, which are not present in \cite{LevineStuart22}.}}
    \label{fig:L63-LS}
\end{figure}

\textcolor{blue-}{The effect of the sampling time $\Delta t$ on the forecasting capability of RFMs has been previously studied by \cite{LevineStuart22}. In particular, they compared a standard RFM with an RFM for which the vector field of the underlying dynamical system is learned.  
The vector field was determined from $\dot{\bb{u}}_n$, which was computed from data $\bb{u}_n$ using splines. In Figure~\ref{fig:L63-LS} we compare their results for $D_r=200$ with our implementation of a standard RFM using a hit-and-run algorithm and with SkipRFM, which as we discussed in Section~\ref{ssec:skip} approximates the tendency via an explicit Euler discretization. To allow for a comparison with the results of \cite{LevineStuart22} we use the validity time $\tau_f$ instead of the VPT, defined by
\begin{align}
\tau_f = \min \left\{ n \Delta t : || \hat{\bb{u}}_{n} - \bb{u}_{n}||_2 \ge  \gamma    \overline{||\bb{u}||}_2 \right\},
\label{eq:tauf}
\end{align}
where the mean $\overline{||\bb{u}||}_2$ is estimated from the training data. We use the same threshold $\gamma=0.05$ as \cite{LevineStuart22}. We show results for several values of the sampling time with 
using a fixed integration time $T=1,000$, which implies that the larger sampling times correspond to smaller amounts of training data. Figure~\ref{fig:L63-LS} illustrates two separate points about RFMs. First, \cite{LevineStuart22} found that for small values of the sampling time $\Delta t$, the RFM which learns the vector field (labelled {\em{rhs} (L+S)}) outperforms the standard RFM (labelled as {\em{RFM (L+S)}}), but this ordering changes for large values of the sampling time. The deterioration of their {\em{rhs (L+S)}} method with a vanishing mean validity time for $\Delta t=0.1$ can be related to the deterioration of the estimate of the time-derivative $\dot{\bb{u}}$ for large sampling times. In contrast, our SkipRFM, which does not learn the vector field but the tendency $\bb{u}_{n+1}-\bb{u}_n$ does not show such deterioration at $\Delta t=0.1$ and never performs worse than the standard RFM. Secondly, \cite{LevineStuart22} uses a Bayesian optimization algorithm to determine all hyperparameters, including the internal weights, whereas we use the hit-and-run Algorithm~\ref{algo:hr}. This leads to a superior performance at large values of $\Delta t$ compared to our standard RFM. However, the hit-and-run algorithm performs significantly better for small sampling times.}

\textcolor{orange-}{The kernel density plots in Figures~\ref{fig:L63-1} and \ref{fig:L63-large-dt} show a large degree of variance of VPT. This is to be expected for an underlying chaotic dynamics, and the distribution of VPT does not significantly change upon increasing the width $D_r$ beyond a certain value. Ideally, one would like to shift the tails of the distribution of VPT towards larger forecast times and avoid occasional short forecast times. In fact, the hit-and-run algorithm achieves this: the presence of features which correspond to the linear and/or saturated region of the activation function, contribute to a higher variance of VPT (see Figure 8 in \cite{mandal2024choice}).}

\begin{figure}[!htp]
    \centering
\includegraphics[width=\linewidth]{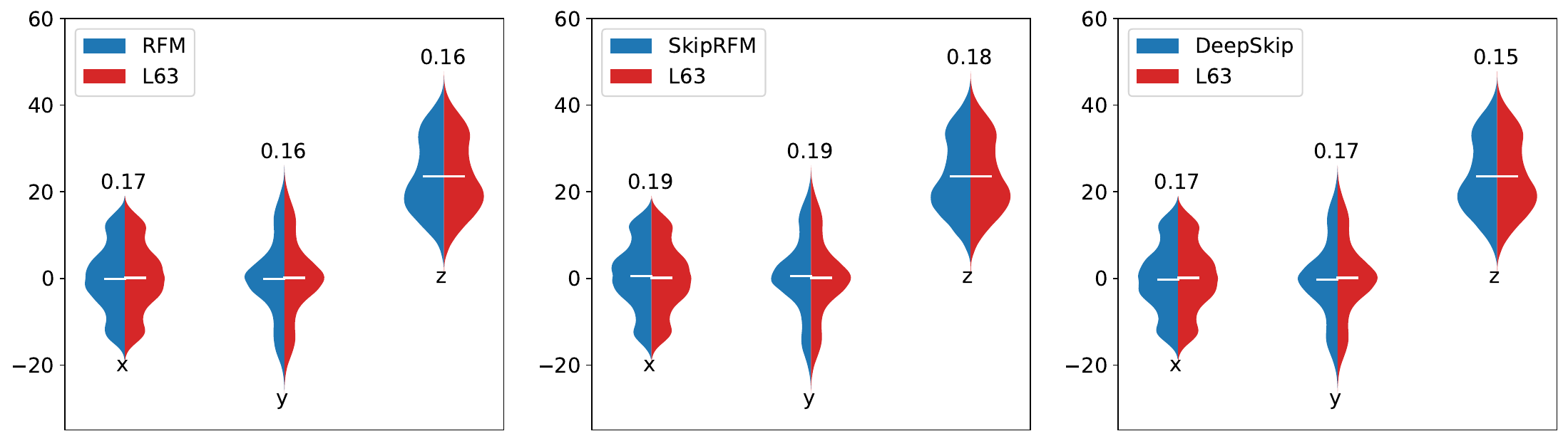}
    \caption{Marginal densities for $x$, $y$ and $z$ estimated from the invariant measures of the L63 system \eqref{eq:L63} and of the surrogate models RFM, SkipRFM and DeepSkip. The respective Wasserstein $W_2$ distances are indicated at the top of the kernel density plots. We used the best models marked with the $*$-symbol in Figure~\ref{fig:L63-1} and shaded in Table~\ref{tab:L63_1_s} to generate the data. \textcolor{orange-}{The mean VPT for these models are $9.8, 10.6$ and $12.0$ from left to right.}}
    \label{fig:L63-climate}
\end{figure}

Besides being able to track individual trajectories, surrogate models need to produce reliable long-term predictions of the statistical features of the underlying dynamical system. Figure~\ref{fig:L63-climate} compares the marginal densities estimated from the invariant measures of the original L63 system \eqref{eq:L63} and the learned surrogate models. The data shown were generated with long simulations spanning $910$ Lyapunov time units. All three surrogate models are able to reproduce the long-term statistics of the L63 system equally well with comparable Wasserstein distances.

\subsection{Lorenz-96}
\label{ssec:L96} 

In this section we demonstrate the forecast skill for the $40$-dimensional Lorenz-96 (L96) system 
\begin{align}
    \frac{dx_i}{dt} = (x_{i+1} - x_{i-2}) x_{i-1} - x_i + F,\;\;i=1,2,\cdots,D,
    \label{eq:L96}
\end{align}
with dimension $D=40$, forcing $F=10$ and periodic boundary conditions $x_{i+D}=x_i$ \cite{lorenz1996predictability}. The maximal Lyapunov exponent is estimated to be $\Lambda = 2.27$ \cite{vlachas2020backpropagation}. We consider here SkipRFM and DeepSkip, and their localized counterparts LocalSkip and LocalDeepSkip. We do not show results for RFM and LocalRFM as their performance is comparable to SkipRFM and LocalSkip. We will see that for this $40$-dimensional dynamical system localization is the dominant factor in ensuring good performance. 

\begin{figure}[!htp]
    \centering
\includegraphics[width=\linewidth]{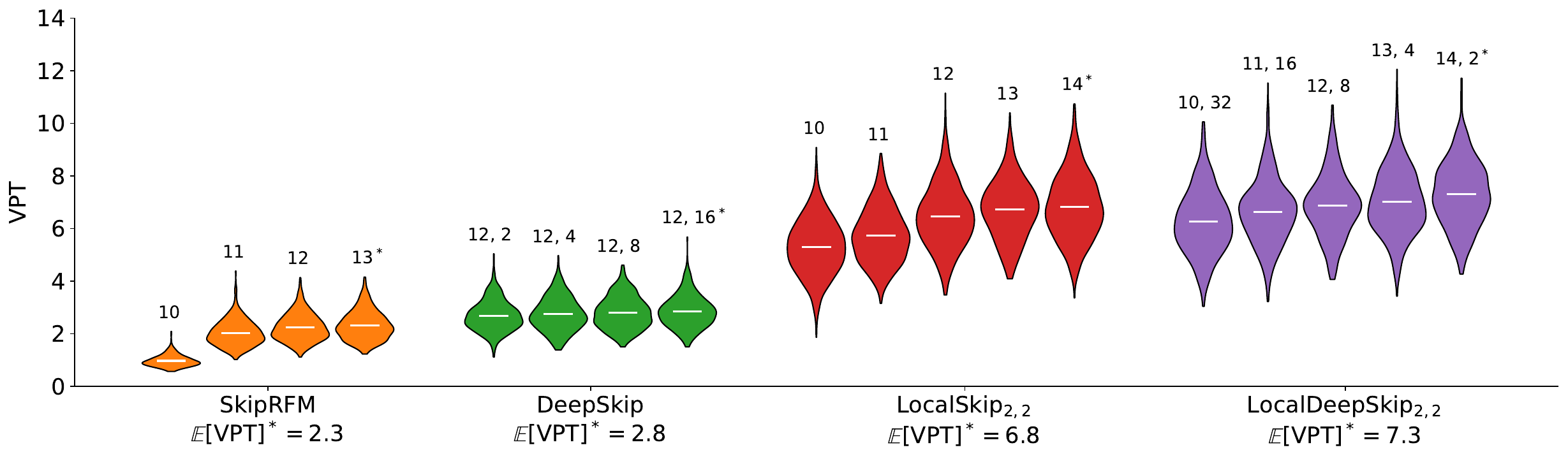}
    \caption{Kernel density plots of VPT for the L96 system \eqref{eq:L96} for $(N, \Delta t, \varepsilon)=(10^5, 0.01, 0.5)$. For shallow variants $\log_2(D_r)$ is indicated on the top of the plots. For deep variants $(\log_2(D_r), B)$ is indicated on the top of each plot. The $*$-symbol indicates the model with the best mean VPT within each architecture.}
    \label{fig:L96-1}
\end{figure}

\begin{figure}
    \centering
    \includegraphics[scale=0.5]{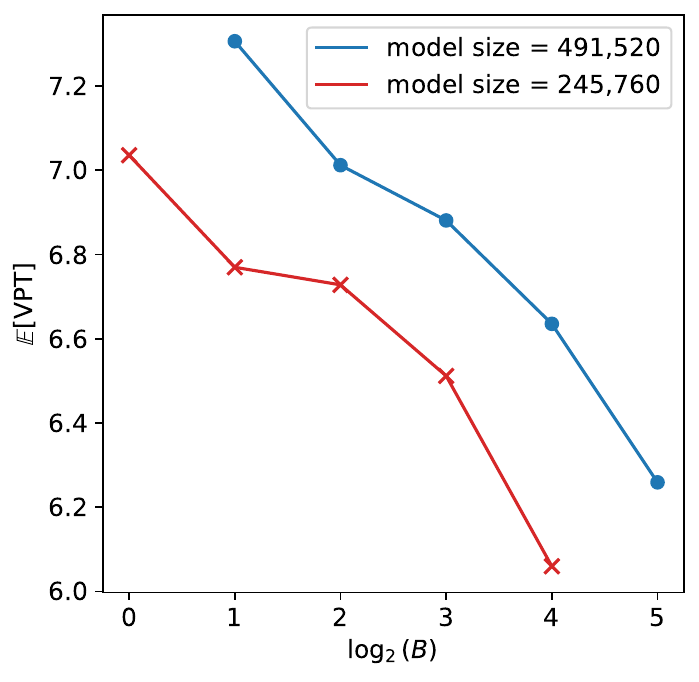}
    \caption{Mean VPT for LocalDeepSkip as a function of depth $B$ for constant model size $S$ and localization scheme $(G,I)=(2,2)$ for the L96 system \eqref{eq:L96}. Along each curve the model size $S=(\hat D+G+1)D_rB$ with $\hat D = 2G(I+1)$, remains constant and the width $D_r$ decreases with depth. }
    \label{fig:L96-depth}
\end{figure}

Figure~\ref{fig:L96-1} shows the distribution of VPT for these models for $N=10^5$, $\Delta t=0.01$ and $\varepsilon=0.5$. We choose a localization scheme with $(G,I)=(2,2)$; see Appendix~\ref{ssec:loc} for different localization schemes and general guidelines for selecting an optimal localization scheme. The positive effect of localization is clearly seen with roughly $3$-times longer forecasting times when compared to the respective non-localized versions. We achieve an optimal mean VPT with $\mathbb{E}[{\rm VPT}] = 7.3$ for LocalDeepSkip with $D_r=16,384$ and $B=2$. The largest localized models that we could accommodate on the GPU were twice as wide as the largest non-localized models, allowing for a much greater expressivity. The performance of non-localized  models plateau quickly with increasing model size whereas we run out of GPU memory before observing saturation in the forecast skill of the localized models. The best deep models are able to forecast approximately $0.5$ Lyapunov time units longer than their shallow counterparts for both localized and non-localized models. Unlike for the L63 system, the performance of deep models decreases with increasing depth when the model size $S=(\hat D+G+1)D_rB$ with $\hat D = 2G(I+1)$ is kept fixed, as seen in Figure~\ref{fig:L96-depth} for two different model sizes.

We see that shallow but wide models perform better than deeper models of the same size by approximately $1$ Lyapunov time unit. We believe that this is due to the more complex nature of the L96 system. The learning task requires (for given data length $N$) a sufficiently large internal layer width $D_r$ to ensure reliable forecasting at each of the $B$ layers of a deep architecture. Since increasing the depth $B$ implies a decrease in the width $D_r$, the deeper networks are not able to resolve the dynamics to sufficient accuracy at each layer. \textcolor{red-}{Hence, deeper architectures with $B>1$ are only beneficial once the width $D_r$ is sufficiently large such that the forecast skill has saturated. Appendix~\ref{ssec:BDr} explores the interplay between the width $D_r$ and the depth $B$ for the L63 system and the L96 system supporting this claim.}

Table~\ref{tab:L96-lit} shows a comparison of our best performing LocalDeepSkip model with recent benchmark results, highlighting that LocalDeepSkip achieves state-of-the-art forecast times with $\mathbb{E}[{\rm VPT}]=7.3$ at $1.3$ orders of magnitude smaller model size. In Appendix~\ref{ssec:detail} a comparison of our variants of the random feature model is shown in Tables~\ref{tab:L96_1_s} and \ref{tab:L96_1_s-l} for different model sizes, reporting on the mean, median, standard deviation of the VPT as well as the maximal and minimal values. 

\begin{table}[!htp]
    \centering
    \begin{tabular}{|c|c|c|c|c|c|c|}\hline
    Source & Model & $\log_{10}(\text{model size})$ & $\mathbb{E}[{\rm VPT}]$ & $N$ & $\Delta t$ & $\varepsilon$\\  \hline
    Penny et al. (2022)~\cite{penny2022integrating, platt2022systematic}\footnote{We remark that the results for Penny et al. were reported in Figure~14 and Table~13 of \cite{platt2022systematic} rather than in \cite{penny2022integrating}}.  
     & RC & $7.56$ & $2.5-2.8$ & $2\times10^5$ & $0.01$ & $0.5$\\ \hline
    Vlachas et al. (2022)~\cite{vlachas2020backpropagation} & Localized LSTM & $5.95$ & $3.9$ & $10^5$ & $0.01$ & $0.5$\\ \hline
         Platt et al. (2022)~\cite{platt2022systematic} & Localized RC & $7.03$ & $6.5-6.8$ & $4\times10^4$ & $0.01$ & $0.5$\\ \hline
        Our work & LocalDeepSkip RFM  & \cellcolor{pink}$5.69$ & \cellcolor{pink}$7.3$ & $10^5$ & $0.01$ & $0.5$\\ \hline
    \end{tabular}
    \caption{Comparison of mean VPT and corresponding model sizes from recent benchmark results for forecasting the L96 system \eqref{eq:L96}. The result corresponding to the best mean VPT $\mathbb{E}[{\rm VPT}]$ is reported for each source. The largest $\mathbb{E}[{\rm VPT}]$ and the corresponding smallest model size are highlighted by red  shading. }
    \label{tab:L96-lit}
\end{table}

\begin{figure}[!htp]
    \centering
\includegraphics[scale=0.4]{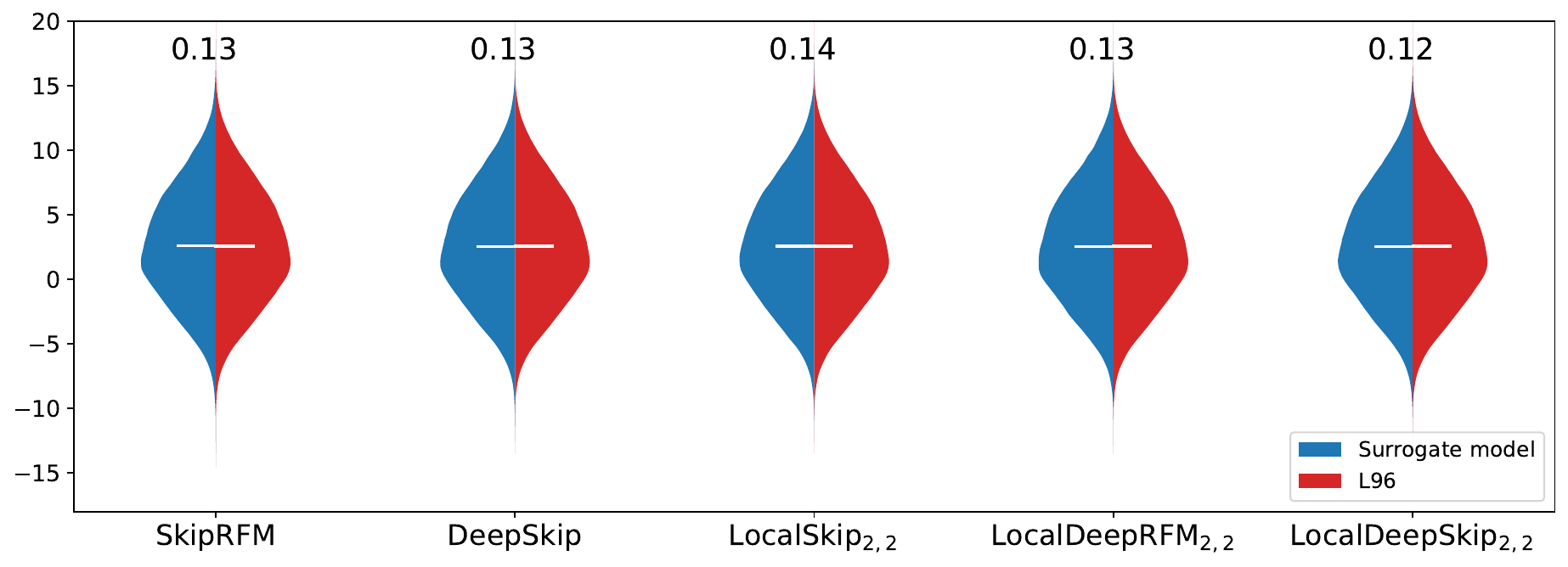}
    \caption{Marginal densities for a component of the L96 system, estimated from the invariant measures of the original L96 system \eqref{eq:L96} and of various surrogate models. We employ translational symmetry and use all $40$ components to estimate the densities. The respective Wasserstein $W_2$ distances are indicated at the top of the kernel density plots. We used the best models marked with the $*$-symbol in Figure~\ref{fig:L96-1} and shaded in Tables~\ref{tab:L96_1_s}, \ref{tab:L96_1_s-l} to generate the data. \textcolor{orange-}{The mean VPT for these models are $2.3, 2.8, 6.8, 7.2$ and $7.3$ from left to right.}}
    \label{fig:L96-climate}
\end{figure}

Figure~\ref{fig:L96-climate} compares the empirical marginal densities estimated from the invariant measures of the original L96 system \eqref{eq:L96} and the learned surrogate models. Both the non-localized and the localized variants are able to reproduce the long-term statistics of the L96 system equally well with comparable Wasserstein distances. 

\subsection{Kuramoto-Sivashinsky}
\label{ssec:KS} 

We further consider the Kuramoto-Sivashinsky (KS) equation 
\begin{align}
    &\frac{\partial u}{\partial t} + u \frac{\partial u}{\partial x} + \frac{\partial^2 u}{\partial x^2} + \frac{\partial^4 u}{\partial x^4} = 0 
\label{eq:KS}
\end{align}
for $x\in[0, L]$ with periodic boundary conditions $u(0, t)=u(L, t)$ as an example of a partial differential equation exhibiting spatio-temporal chaos \cite{kuramoto1978diffusion,ashinsky1988nonlinear}. The maximal Lyapunov exponent is estimated to be $\Lambda = 0.094$ \cite{vlachas2020backpropagation}. We solve equation~\ref{eq:KS} on a domain of length $L=200$ with a uniform grid of $512$ nodes using the ETDRK4 method \cite{kassam2005fourth} with a time step of $h=0.001$. The data are subsampled in time to produce a time series of $512$-dimensional states of length $N=10^5$ with sample time $\Delta t=0.25$ for the learning task. We employ a VPT error threshold of $\varepsilon=0.5$. For this high-dimensional system localization is essential to obtain any reliable forecasting skill. We choose a localization scheme of $(G,I)=(8,1)$; see Appendix~\ref{ssec:loc} for different localization schemes and general guidelines for selecting an optimal localization scheme. To allow for a large expressive model capable of capturing the complexity of the chaotic dynamics, we focus mainly on deep architectures. 

\begin{figure}
    \centering
    \includegraphics[width=\linewidth]{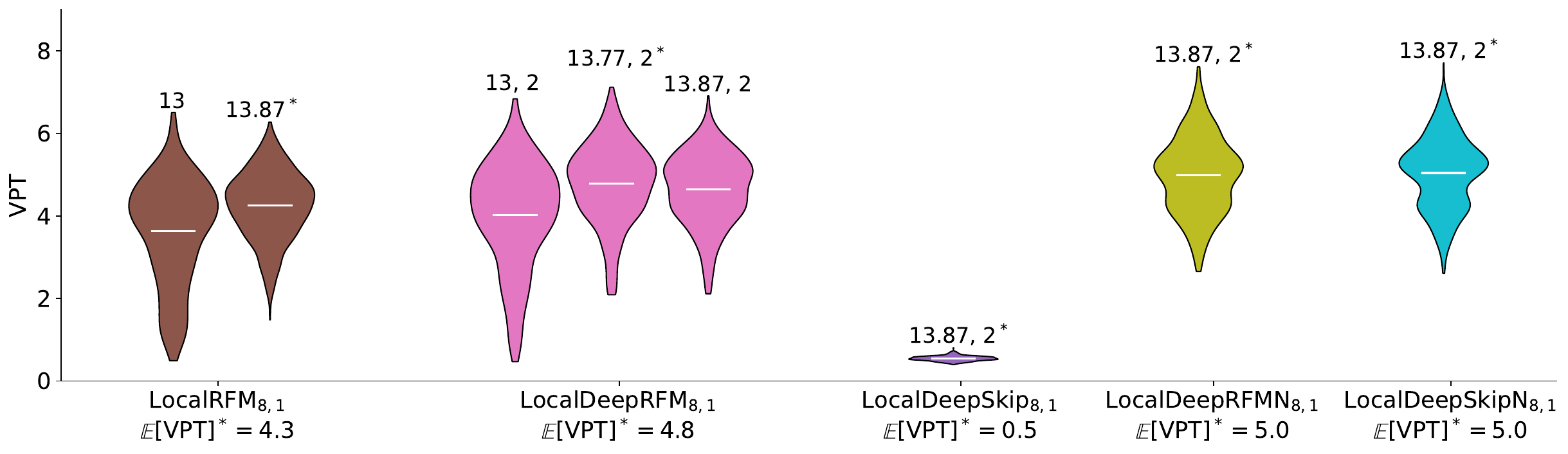}
    \caption{Kernel density plots of VPT for the KS system \eqref{eq:KS} for $(N, \Delta t, \varepsilon)=(10^5, 0.25, 0.5)$ for various localized surrogate models. For all models $(\log_2(D_r), B)$ is indicated on the top of each plot. The $*$-symbol indicates the model with the best mean VPT within each architecture. The maximal value of $D_r$ allowed by our GPU is $15,000$ (i.e. $\log_2(D_r)\approx 13.87$).}
    \label{fig:KS_1}
\end{figure}

The trajectory data for KS generated by the ETDRK4 algorithm has a large condition number $\sim10^{15}$. As discussed in Section~\ref{ssec:noise}, ill-conditioned data matrices imply ill-conditioned learned outer weight matrices $\bb{W}$ which has a catastrophic effect on long-term forecasts and might also affect short-term forecasts. The LocalDeepRFM and LocalDeepSkip models tested on this problem have outer weight matrices with condition numbers $\sim950$ and $\sim1,350$ respectively. Due to the larger condition number, LocalDeepSkip performs much worse than LocalDeepRFM, with $\mathbb{E}[{\rm VPT}]=0.5$ for LocalDeepSkip and $\mathbb{E}[{\rm VPT}]=4.8$ for LocalDeepRFM (cf. Figure~\ref{fig:KS_1}). This is contrary to our observations that skip connections improve performance for the L63 and the L96 system. We remark that LocalSkip models perform equally badly when trained on ill-conditioned data. A possible reason for the high condition numbers of the $\bb{W}$ matrix for skip connections may be the following. For skip connections $\bb{W}$ depends on the matrix of differences $\bb{u}_{n+1}-\bb{u}_{n}$ rather than just on $\bb{u}_{n+1}$. Hence, its condition number depends on the condition number of this difference matrix. For the KS equation significant values of the differences $\bb{u}_{n+1}-\bb{u}_{n} \in \mathbb{R}^{512}$ appear only in small spatially localized regions with small entries in most components, implying a large condition number.  

To mitigate the detrimental effect of large condition numbers, we add zero-mean Gaussian noise with standard deviation $10^{-3}$ to the training data. LocalDeepSkip models trained on artificially noisy data are able forecast up to $5$ Lyapunov units on average, as shown in Figure~\ref{fig:KS_1}. Models with and without skip connections are seen to perform equally well when the training data are artificially contaminated by small but non-negligible noise. 

A comparison of our results with the benchmark results of \cite{vlachas2020backpropagation}, where the same experimental setup was used, is reported in Table~\ref{tab:KS-lit} and shows that LocalDeepSkip trained on artificially noised data achieves marginally better results with a mean VPT of $\mathbb{E}[{\rm VPT}]=5.0$ but with an approximately $2.7$ orders of magnitude smaller model. In Appendix~\ref{ssec:detail} a comparison of our variants of the random feature model is shown in Table~\ref{tab:KS_1} for different model sizes, reporting on the mean, median, standard deviation of the VPT as well as the maximal and minimal values. 

\begin{table}[!htp]
    \centering
    \begin{tabular}{|c|c|c|c|c|c|c|}\hline
    Source & Model & $\log_{10}(\text{model size})$ & $\mathbb{E}[{\rm VPT}]$ & $N$ & $\Delta t$ & $\varepsilon$ \\  \hline
    Vlachas et al. (2022)~\cite{vlachas2020backpropagation} & Localized RC & $8.77$ & $4.8$ & $10^5$ & $0.25$ & $0.5$ \\ \hline
    Our work  & LocalDeepSkipN RFM & \cellcolor{pink}$6.09$ & \cellcolor{pink}$5.0$ & $10^5$ & $0.25$ & $0.5$ \\ \hline        
    \end{tabular}
    \caption{Comparison of mean VPT and corresponding model sizes from recent benchmark results for forecasting the KS system \eqref{eq:KS} with $512$ spatial grid points on a domain of length $L=200$. The result corresponding to the best mean VPT $\mathbb{E}[{\rm VPT}]$ is reported for each source. The largest $\mathbb{E}[{\rm VPT}]$ and the corresponding smallest model size are indicated with coloring.}
    \label{tab:KS-lit}
\end{table}

\begin{figure}[!htp]
    \centering
    \includegraphics[scale=0.5]{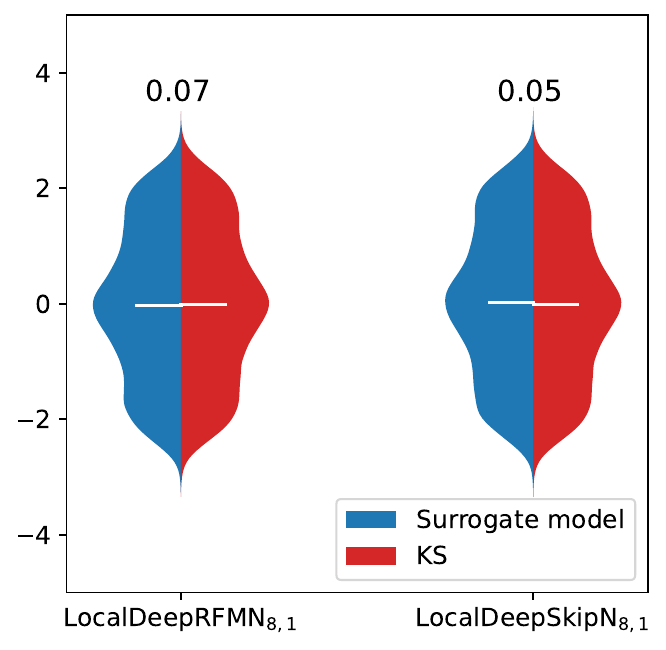}
    \caption{Marginal densities for a component of the invariant measure of the KS system, estimated from the original KS system \eqref{eq:KS} and from various surrogate models. We employ translational symmetry and use all $512$ components to estimate the densities. The respective Wasserstein $W_2$ distances are indicated at the top of the kernel density plots. We used the best models marked with the $*$-symbol in Figure~\ref{fig:KS_1} and shaded in Table~\ref{tab:KS_1} to generate the data. \textcolor{orange-}{The mean VPT for both of these models is $5.0$.}}
    \label{fig:KS-climate}
\end{figure}

For reproducing long-term statistics, models trained on  noise-free ill-conditioned data are not suitable since they accumulate large errors during long simulations. However, models trained on noisy well-conditioned data are able to reproduce the invariant measure of the KS equation, as seen in Figure~\ref{fig:KS-climate}. The LocalDeepRFM and the LocalDeepSkip architectures with artificially added noise are able to reproduce the long-term statistics of the KS system equally well with comparable Wasserstein distances. We remark that LocalDeepRFM trained on pure data or on noisy data performs approximately equally well on short time scales. However, without the addition of artificial noise to the training data, all surrogate models exhibit numerical instability for long-time forecasting.  

\section{Discussion}
\label{sec:conclude}

In this work we extend random feature maps with a $\tanh$-activation function by introducing skip connections, a deep architecture and localization with the aim to produce reliable surrogate models for dynamical systems. We considered a $3$-dimensional Lorenz-63 system, a $40$-dimensional Lorenz-96 system and a $512$-dimensional finite difference discretization of the Kuramoto-Sivashinsky equation, and studied the ability of the learned surrogate models to forecast individual trajectories as well as the long-time statistical behaviour. In all three systems our modifications lead to either better or equal performance when compared to recent benchmark results using RCs or LSTMs, with orders of magnitude smaller models. For all architectures we judiciously chose the internal weights using the computationally efficient hit-and-run algorithm developed in \cite{mandal2024choice}. This algorithm ensured that for the training data the features were neither linear nor saturated and took advantage of the nonlinear nature of the $\tanh$-activation function.

We showed in which situations each of our modifications can be beneficial and that they can significantly improve the forecast capabilities of random feature models. We showed that introducing skip connections typically leads to better performance. However, when the data matrix has too high a condition number, ridge regression leads to an ill-conditioned trained outer weight matrix. This renders the learned surrogate model unstable and unreliable as a forecast model. To combat this, we proposed to add small artificial noise to the data. This allowed for state-of-the-art forecast times for Kuramoto-Sivashinsky equation with more than an order of magnitude smaller model size when compared against recent benchmark results.  \textcolor{blue-}{It is well known that adding sufficiently strong noise to the training data can severely affect the training of RFMs (see, for example, \cite{GottwaldReich21}). We only apply very small noise such that by eye the training data appear unchanged. Although adding noise to the training data is frequently used  \cite{VlachasEtAl18, vlachas2020backpropagation}, it may be beneficial to add the noise on the features $\mathbf{\Phi}$ instead.}

For higher dimensional systems localization was found to be essential. The optimal choice of the localization scheme requires balancing the required accuracy for a given data set of length $N$, the decay of the spatial correlations of the underlying dynamical system and the available GPU memory.  

Our simulations suggest that the performance of random feature models can be significantly improved by considering a deep architecture chaining RFM units together where each unit is individually trained to match the data. However, the improvement can only be observed once the width of each individual layer is large enough to allow for a sufficiently accurate representation of the dynamics. For instance, for the Lorenz-96 system, we observed that the localized models had not yet plateaued with increasing $D_r$, and the available GPU memory was fully utilized before reaching saturation.

Our random feature map variants can achieve comparable or even superior performance to RCs while requiring only a fraction of the model size and hence, computational effort. Moreover, although RFMs and RCs share similar learning mechanisms, RFMs offer several advantages over their RC counterparts. One key advantage is that RCs require tuning multiple hyperparameters, such as the spectral radius and density of the reservoir adjacency matrix, degrees of freedom, leak rate, strength of the input signal, strength of the input bias, regularization etc \cite{platt2022systematic}, which is computationally expensive. In contrast, RFMs only require optimization of the regularization hyperparameter. Furthermore, RCs are comprised of layers similar to RFMs and a reservoir. These reservoirs are represented by weight matrices of size $D_r^2$ whereas the weight matrices in RFMs have size $DD_r$. Since typically, $D_r\gg D$, for the same width, RFMs are significantly lighter models compared to RCs. We remark that implementing sparse matrix and dense vector multiplication on a GPU is not efficient unless the matrix is very sparse. However, the reservoirs employed in the benchmark results reported here are not sparse e.g. \cite{platt2022systematic} reports the density of the RC adjacency matrix as being $0.98$. To deal with the high memory demands for large RC models a batched approach was used in \cite{vlachas2020backpropagation}. 

\textcolor{blue-}{Recently, Bayesian methods were proposed to estimate the RFM hyperparameters, including the internal weights \cite{LevineStuart22,DunbarEtAl24}. It appears that Bayesian hyperparameter tuning has advantages for large sampling times, whereas the hit-and-run sampling algorithm seems to be beneficial for smaller sampling times (cf. Figure~\ref{fig:L63-LS}). Combining these two approaches could further improve the forecasting capabilities of RFMs. The Bayesian optimization strategy can also be employed to more efficiently tune hyperparameters such as the regularization hyperparameter which was determined here using grid-search and the localization scheme.}

\textcolor{blue-}{Our skip connection is of the form of a forward-Euler numerical integrator for an underlying continuous time dynamical system. One could aim to learn higher-order multistep integrators such as the Runge-Kutta integrator to improve the accuracy of the prediction as studied in \cite{KellerDu21,DuEtAl22}. However, one needs to be wary of potential "inverse crimes" where the learned map is used for forecasting with a time step different to the sampling time $\Delta t$ used for training, which may result in numerical instabilities \cite{KrishnapriyanEtAl23}.}

We considered here noise-free and complete observations \textcolor{red-}{for a set dynamical systems with known equations, allowing for benchmarking. Data from real-world systems typically are noise-contaminated and the system is accessible only via partial observations. It will be interesting to see if the superior forecasting skill in the case of noise-free and complete observations extends to this relevant case. There has been recent progress on learning real-world dynamical systems using Bayesian learning methods with remarkable accuracy such as {\em{eSPA}} \cite{Horenko22,HorenkoEtAl23,GroomEtAl24} and {\em{BayesNF}} \cite{SaadEtAl24} which provide benchmarks to test against.} \textcolor{red-}{The forecasting capability of RFMs quickly deteriorates when observations are contaminated by noise. However, when combining RFMs with data assimilation procedures such as the ensemble Kalman filter, the noise can be successfully controlled for training RFMs~\cite{gottwald2021supervised}. The lack of complete observations renders the dynamical system for the observed states non-Markovian. A Markovian dynamical system can be achieved by formulating the learning task in an enlarged space of time-delay coordinates \cite{Takens81}. This requires determining an appropriate delay embedding \cite{KantzSchreiber}. These techniques have been shown to be applicable for learning RFMs from partial observations \cite{GottwaldReich21}. Further, to improve the forecast skill of RFMs in the relevant case of partial noisy observations, it may be beneficial to combine our modifications of RFMs with the hybrid approach promoted by \cite{LevineStuart22}. This is planned for further research.}

\section{Acknowledgments} The authors acknowledge support from the Australian Research Council under Grant No. DP220100931.

\section{Author contributions}
PM implemented the random feature models and ran the simulations. PM and GAG equally contributed in conceptualizing the methodology and writing the manuscript.
\section{Competing Interests} The authors declare that they have no competing interests.
\bibliographystyle{naturemag}
\bibliography{references}

\newpage
\section{Appendix}
\label{sec:appendix}


\subsection{One-shot hit-and-run algorithm to draw internal weights for random feature maps}
\label{sec:app:algo:hr}
$\;\;$
\begin{algorithm}[!htp]
\caption{Hit-and-run sampling for a row of the internal augmented matrix $\bb{W}_{\rm in}|\bb{b}_{\rm in}$}
\label{algo:hr}
\begin{algorithmic}[1]
\STATE Input: data $\bb{U}=[\bb{u_1}, \bb{u}_2, \cdots, \bb{u}_N]$. Boundaries for the good range of the $\tanh$-function $L_{0,1}$. Here $L_0=0.4$ and $L_1=3.5$.
\STATE Sample $b$ uniformly from $(L_0, L_1)$, the "good" part of the domain of $\tanh$.
\STATE Select a sign vector $\bb{s}$ uniformly randomly from $\{-1, 1\}^D$.
\FOR{$i = 1,\cdots,D$}
    \IF{$\bb{s}_i = 1$}
        \STATE $\bb{x}_{-,i} \leftarrow \underset{1\le n\le N}{\min} \bb{u}_{n,i}$
        \STATE $\bb{x}_{+,i} \leftarrow \underset{1\le n\le N}{\max} \bb{u}_{n,i}$
    \ELSE
        \STATE $\bb{x}_{-,i} \leftarrow \underset{1\le n\le N}{\max} \bb{u}_{n,i}$
        \STATE $\bb{x}_{+,i} \leftarrow \underset{1\le n\le N}{\min} \bb{u}_{n,i}$
    \ENDIF
\ENDFOR
\STATE $V\leftarrow\{\bb{w}\in\mathbb{R}^D:\sgn(\mathbf{w}_i)\in\{\mathbf{s}_i, 0\}\;\;\forall\;i=1,2,\ldots,D\}$
\STATE Randomly select a unit vector $\bb{d}\in V$. 
    \STATE $c_0\leftarrow0$.
    \STATE $c_1\leftarrow\inf\left(\left\{\frac{L_0-b}{\bb{d}\cdot \bb{x}_-},\frac{L_1-b}{\bb{d}\cdot \bb{x}_+}\right\}\cap(\mathbb R_{>0}\cup\{+\infty\})\right)$ with the convention $\inf\varnothing=+\infty$.
    \STATE Sample $c$ uniformly from $(c_0, c_1)$.
    \STATE $\bb{w}\leftarrow c\bb{d}$
\STATE Uniformly sample a scalar $z$ from $\{-1, 1\}$.
\IF{$z=1$}
    \STATE $(\bb{w},b)$ is our final row sample. 
\ELSE
    \STATE $-(\bb{w},b)$ is our final row sample. 
\ENDIF
\end{algorithmic}
\end{algorithm}


\subsection{Additional numerical details}
\label{ssec:detail}

In this section we document additional numerical details corresponding to the experimental results in Section~\ref{sec:results} for the Lorenz-63 system, the Lorenz-96 system and the Kuramoto-Sivashinsky equation. We present tables summarizing our results. Each row summarizes the results for $500$ samples that differ in their training data, testing data, and the non-trainable random weights and biases of the corresponding model. Along with the model details and mean, standard deviation, median, minimum and maximum of VPT, each row also shows the corresponding value of the regularization hyperparameter $\beta$ used in the experiments as well as the average training time in seconds which includes the run-time of algorithm~\ref{algo:hr}. For all models trained on noisy data, a zero-mean Gaussian noise with standard deviation $10^{-3}$ was used. For each architecture, the best performing model has been highlighted by a red shading.


\subsubsection{Lorenz-63}
\label{ssec:ap-L63} 

We use two different setups for the L63 system. Table~\ref{tab:L63_1_s} documents results for $(N, \Delta t, \varepsilon) = (5\times10^4, 0.01, 0.3)$ which is also used in \cite{platt2022systematic}. Table~\ref{tab:L63_0_s} documents results for $(N, \Delta t, \varepsilon) = (2\times10^4, 0.02, \sqrt{0.05})$ corresponding to the setup used in \cite{gottwald2021supervised, mandal2024choice}. A similar setup with $\varepsilon=0.4$ appears in \cite{akiyama2022computational}. To generate the training and testing data for L63, we use a burn-in period of $40$ model time units. 

\begin{table}[!htp]
    \centering
    \begin{tabular}{|c|c|c|c|c|c|c|c|c|c|c|} \hline
\multicolumn{4}{|c|}{Model} &\multicolumn{5}{c|}{VPT} & \multicolumn{2}{c|}{}\\ \hline
architecture & $D_r$ & $B$ & model size & mean & std & median & min & max &$\beta$ & $\mathbb{E}[t_{\rm train}]$(s)\\ \hline\hline
\multirow{6}{*}{RFM} & \cellcolor{pink}512 & \cellcolor{pink}1 & \cellcolor{pink}3,584 & \cellcolor{pink}9.8 & \cellcolor{pink}1.8 & \cellcolor{pink}9.8 & \cellcolor{pink}4.0 & \cellcolor{pink}16.0 & \cellcolor{pink}3.52e-09 & \cellcolor{pink}1.1e-02\\ \cline{2-11}
 & 1,024 & 1 & 7,168 & 9.8 & 1.5 & 9.8 & 4.9 & 15.5 & 6.40e-09 & 1.6e-02\\ \cline{2-11}
 & 2,048 & 1 & 14,336 & 9.3 & 1.5 & 9.3 & 4.0 & 15.3 & 4.96e-08 & 4.4e-02\\ \cline{2-11}
 & 4,096 & 1 & 28,672 & 9.3 & 1.5 & 9.3 & 3.8 & 13.8 & 8.20e-08 & 1.2e-01\\ \cline{2-11}
 & 8,192 & 1 & 57,344 & 9.5 & 1.7 & 9.5 & 3.8 & 16.3 & 6.76e-08 & 4.4e-01\\ \cline{2-11}
 & 16,384 & 1 & 114,688 & 9.5 & 1.5 & 9.6 & 4.9 & 18.9 & 8.92e-08 & 2.1e+00\\ \cline{2-11}
\hline\hline
\multirow{6}{*}{SkipRFM} & 512 & 1 & 3,584 & 10.1 & 1.7 & 10.0 & 4.1 & 16.9 & 3.88e-09 & 7.2e-03\\ \cline{2-11}
 & 1,024 & 1 & 7,168 & 10.5 & 1.5 & 10.5 & 5.5 & 15.1 & 6.40e-09 & 1.6e-02\\ \cline{2-11}
 & 2,048 & 1 & 14,336 & 10.3 & 1.6 & 10.3 & 5.0 & 19.5 & 3.16e-08 & 4.4e-02\\ \cline{2-11}
 & 4,096 & 1 & 28,672 & 10.3 & 1.5 & 10.3 & 6.5 & 16.2 & 7.12e-08 & 1.2e-01\\ \cline{2-11}
 & \cellcolor{pink}8,192 & \cellcolor{pink}1 & \cellcolor{pink}57,344 & \cellcolor{pink}10.6 & \cellcolor{pink}1.5 & \cellcolor{pink}10.7 & \cellcolor{pink}5.2 & \cellcolor{pink}16.1 & \cellcolor{pink}6.76e-08 & \cellcolor{pink}4.4e-01\\ \cline{2-11}
 & 16,384 & 1 & 114,688 & 10.4 & 1.6 & 10.4 & 5.0 & 17.2 & 2.44e-07 & 2.1e+00\\ \cline{2-11}
\hline\hline
\multirow{20}{*}{DeepRFM} & 1024 & 1 & 10240 & 9.7 & 1.6 & 9.7 & 4.0 & 15.0 & 4.96e-09 & 1.8e-02\\ \cline{2-11}
 & \cellcolor{pink} 1024 & \cellcolor{pink} 2 & \cellcolor{pink} 20480 & \cellcolor{pink} 10.0 & \cellcolor{pink} 1.6 & \cellcolor{pink} 9.9 & \cellcolor{pink} 5.5 & \cellcolor{pink} 18.4 & \cellcolor{pink} 4.96e-09 & \cellcolor{pink} 3.2e-02\\ \cline{2-11}
 & 2048 & 1 & 20480 & 9.3 & 1.8 & 9.2 & 3.8 & 16.7 & 3.16e-08 & 4.5e-02\\ \cline{2-11}
 & 1024 & 4 & 40960 & 9.9 & 1.6 & 9.8 & 4.1 & 15.3 & 4.96e-09 & 6.3e-02\\ \cline{2-11}
 & 2048 & 2 & 40960 & 9.5 & 1.6 & 9.3 & 5.0 & 14.9 & 3.16e-08 & 7.5e-02\\ \cline{2-11}
 & 4096 & 1 & 40960 & 9.6 & 1.5 & 9.6 & 4.1 & 14.9 & 5.32e-08 & 1.2e-01\\ \cline{2-11}
 & 1024 & 8 & 81920 & 9.8 & 1.7 & 9.8 & 4.8 & 14.6 & 4.96e-09 & 1.1e-01\\ \cline{2-11}
 & 2048 & 4 & 81920 & 9.4 & 1.6 & 9.3 & 4.1 & 14.9 & 3.16e-08 & 1.5e-01\\ \cline{2-11}
 & 4096 & 2 & 81920 & 9.8 & 1.7 & 9.8 & 4.0 & 19.0 & 5.32e-08 & 2.5e-01\\ \cline{2-11}
 & 8192 & 1 & 81920 & 9.7 & 1.6 & 9.7 & 3.9 & 15.0 & 9.28e-08 & 4.7e-01\\ \cline{2-11}
 & 1024 & 16 & 163840 & 9.7 & 1.6 & 9.8 & 4.1 & 14.5 & 4.96e-09 & 2.4e-01\\ \cline{2-11}
 & 2048 & 8 & 163840 & 9.4 & 1.6 & 9.3 & 3.8 & 16.2 & 3.16e-08 & 3.1e-01\\ \cline{2-11}
 & 4096 & 4 & 163840 & 9.5 & 1.7 & 9.6 & 4.0 & 15.2 & 5.32e-08 & 5.0e-01\\ \cline{2-11}
 & 8192 & 2 & 163840 & 9.8 & 1.7 & 9.8 & 3.9 & 15.4 & 9.28e-08 & 9.4e-01\\ \cline{2-11}
 & 16384 & 1 & 163840 & 9.9 & 1.6 & 9.8 & 3.8 & 16.4 & 8.92e-08 & 2.1e+00\\ \cline{2-11}
 & 1024 & 32 & 327680 & 9.7 & 1.6 & 9.7 & 4.0 & 14.3 & 4.96e-09 & 4.6e-01\\ \cline{2-11}
 & 2048 & 16 & 327680 & 9.4 & 1.5 & 9.4 & 3.9 & 14.8 & 3.16e-08 & 6.4e-01\\ \cline{2-11}
 & 4096 & 8 & 327680 & 9.7 & 1.7 & 9.7 & 3.8 & 17.9 & 5.32e-08 & 1.0e+00\\ \cline{2-11}
 & 8192 & 4 & 327680 & 9.6 & 1.7 & 9.6 & 3.8 & 15.7 & 9.28e-08 & 1.9e+00\\ \cline{2-11}
 & 16384 & 2 & 327680 & 9.9 & 1.6 & 9.9 & 3.8 & 15.4 & 8.92e-08 & 4.3e+00\\ \cline{2-11}
\hline\hline
\multirow{17}{*}{DeepSkip} & 1,024 & 1 & 10,240 & 10.1 & 1.7 & 10.0 & 4.0 & 17.0 & 4.96e-09 & 1.6e-02\\ \cline{2-11}
 & 1,024 & 2 & 20,480 & 10.9 & 1.7 & 11.0 & 4.9 & 17.7 & 4.96e-09 & 3.2e-02\\ \cline{2-11}
 & 1,024 & 4 & 40,960 & 11.3 & 1.7 & 11.3 & 4.9 & 18.4 & 4.96e-09 & 6.2e-02\\ \cline{2-11}
 & 4,096 & 1 & 40,960 & 9.9 & 1.6 & 9.9 & 3.9 & 18.6 & 5.32e-08 & 1.2e-01\\ \cline{2-11}
 & 1,024 & 8 & 81,920 & 11.4 & 1.6 & 11.5 & 5.3 & 16.6 & 4.96e-09 & 1.2e-01\\ \cline{2-11}
 & 4,096 & 2 & 81,920 & 10.9 & 1.7 & 11.0 & 4.9 & 19.5 & 5.32e-08 & 2.4e-01\\ \cline{2-11}
 & 8,192 & 1 & 81,920 & 10.3 & 1.6 & 10.3 & 4.9 & 16.8 & 6.76e-08 & 4.4e-01\\ \cline{2-11}
 & 1,024 & 16 & 163,840 & 11.7 & 1.7 & 11.8 & 6.3 & 21.2 & 4.96e-09 & 2.4e-01\\ \cline{2-11}
 & 2,048 & 8 & 163,840 & 11.4 & 1.7 & 11.4 & 5.7 & 16.7 & 2.19e-08 & 3.2e-01\\ \cline{2-11}
 & 4,096 & 4 & 163,840 & 11.0 & 1.6 & 11.1 & 4.9 & 16.4 & 5.32e-08 & 4.9e-01\\ \cline{2-11}
 & 8,192 & 2 & 163,840 & 11.0 & 1.5 & 11.1 & 5.0 & 16.0 & 6.76e-08 & 9.0e-01\\ \cline{2-11}
 & 16,384 & 1 & 163,840 & 10.7 & 1.6 & 10.7 & 5.6 & 17.8 & 8.92e-08 & 2.1e+00\\ \cline{2-11}
 & \cellcolor{pink} 1,024 & \cellcolor{pink} 32 & \cellcolor{pink} 327,680 & \cellcolor{pink} 12.0 & \cellcolor{pink} 1.5 & \cellcolor{pink} 12.0 & \cellcolor{pink} 6.4 & \cellcolor{pink} 20.1 & \cellcolor{pink} 4.96e-09 & \cellcolor{pink} 4.6e-01\\ \cline{2-11}
 & 2,048 & 16 & 327,680 & 11.5 & 1.6 & 11.4 & 5.7 & 17.5 & 2.19e-08 & 6.4e-01\\ \cline{2-11}
 & 4,096 & 8 & 327,680 & 11.2 & 1.6 & 11.3 & 6.3 & 18.4 & 5.32e-08 & 1.0e+00\\ \cline{2-11}
 & 8,192 & 4 & 327,680 & 11.2 & 1.6 & 11.3 & 5.2 & 18.3 & 6.76e-08 & 1.8e+00\\ \cline{2-11}
 & 16,384 & 2 & 327,680 & 10.9 & 1.5 & 10.9 & 5.0 & 17.3 & 8.92e-08 & 4.3e+00\\ \cline{2-11}
\cline{1-2}
\end{tabular}
    \caption{Results for the L63 system with $N=5\times10^4$, $\Delta t=0.01$ and $\varepsilon=0.3$ for various surrogate models.}
    \label{tab:L63_1_s}
\end{table}

\begin{table}[!htp]
    \centering
    \begin{tabular}{|c|c|c|c|c|c|c|c|c|c|c|} \hline
\multicolumn{4}{|c|}{Model} &\multicolumn{5}{c|}{VPT} & \multicolumn{2}{c|}{}\\ \hline
architecture & $D_r$ & $B$ & model size & mean & std & median & min & max &$\beta$ & $\mathbb{E}[t_{\rm train}]$(s)\\ \hline\hline
\multirow{6}{*}{SkipRFM} & 512 & 1 & 3,584 & 10.1 & 1.7 & 10.0 & 4.6 & 16.7 & 6.04e-10 & 8.8e-03\\ \cline{2-11}
 & \cellcolor{pink}1,024 & \cellcolor{pink}1 & \cellcolor{pink}7,168 & \cellcolor{pink}10.4 & \cellcolor{pink}1.4 & \cellcolor{pink}10.4 & \cellcolor{pink}4.8 & \cellcolor{pink}16.1 & \cellcolor{pink}8.74e-10 & \cellcolor{pink}1.1e-02\\ \cline{2-11}
 & 2,048 & 1 & 14,336 & 10.0 & 1.5 & 10.1 & 4.7 & 17.5 & 4.24e-09 & 2.6e-02\\ \cline{2-11}
 & 4,096 & 1 & 28,672 & 10.1 & 1.4 & 10.1 & 4.8 & 15.7 & 9.46e-09 & 6.2e-02\\ \cline{2-11}
 & 8,192 & 1 & 57,344 & 10.1 & 1.5 & 10.1 & 4.7 & 15.9 & 2.26e-08 & 2.2e-01\\ \cline{2-11}
 & 16,384 & 1 & 114,688 & 10.3 & 1.5 & 10.2 & 4.7 & 15.9 & 2.62e-08 & 8.9e-01\\ \cline{2-11}
\hline\hline
\multirow{15}{*}{DeepSkip} & 1,024 & 1 & 10,240 & 9.7 & 1.6 & 9.7 & 4.6 & 16.0 & 9.46e-10 & 1.1e-02\\ \cline{2-11}
 & 1,024 & 2 & 20,480 & 10.9 & 1.6 & 10.7 & 4.8 & 17.3 & 9.46e-10 & 2.1e-02\\ \cline{2-11}
 & 1,024 & 4 & 40,960 & 11.0 & 1.6 & 10.8 & 5.6 & 17.7 & 9.46e-10 & 4.2e-02\\ \cline{2-11}
 & 4,096 & 1 & 40,960 & 9.8 & 1.6 & 9.8 & 4.6 & 16.9 & 9.28e-09 & 6.2e-02\\ \cline{2-11}
 & 1,024 & 8 & 81,920 & 11.3 & 1.5 & 11.2 & 4.8 & 17.5 & 9.46e-10 & 7.7e-02\\ \cline{2-11}
 & 4,096 & 2 & 81,920 & 10.6 & 1.5 & 10.6 & 4.7 & 17.1 & 9.28e-09 & 1.3e-01\\ \cline{2-11}
 & 8,192 & 1 & 81,920 & 9.2 & 1.4 & 9.2 & 4.6 & 14.2 & 3.70e-08 & 2.2e-01\\ \cline{2-11}
 & 1,024 & 16 & 163,840 & 11.7 & 1.6 & 11.6 & 6.4 & 18.2 & 9.46e-10 & 1.5e-01\\ \cline{2-11}
 & 4,096 & 4 & 163,840 & 10.8 & 1.6 & 10.7 & 4.8 & 18.2 & 9.28e-09 & 2.5e-01\\ \cline{2-11}
 & 8,192 & 2 & 163,840 & 10.4 & 1.5 & 10.4 & 5.4 & 17.8 & 3.70e-08 & 4.4e-01\\ \cline{2-11}
 & 16,384 & 1 & 163,840 & 9.5 & 1.4 & 9.6 & 4.7 & 14.5 & 5.14e-08 & 8.9e-01\\ \cline{2-11}
 & \cellcolor{pink}1,024 & \cellcolor{pink}32 & \cellcolor{pink}327,680 & \cellcolor{pink}11.8 & \cellcolor{pink}1.5 & \cellcolor{pink}11.7 & \cellcolor{pink}7.8 & \cellcolor{pink}18.2 & \cellcolor{pink}9.46e-10 & \cellcolor{pink}3.1e-01\\ \cline{2-11}
 & 4,096 & 8 & 327,680 & 11.0 & 1.4 & 11.0 & 5.6 & 18.2 & 9.28e-09 & 5.1e-01\\ \cline{2-11}
 & 8,192 & 4 & 327,680 & 10.6 & 1.5 & 10.6 & 5.5 & 15.2 & 3.70e-08 & 9.0e-01\\ \cline{2-11}
 & 16,384 & 2 & 327,680 & 10.5 & 1.5 & 10.5 & 5.5 & 18.2 & 5.14e-08 & 1.8e+00\\ \cline{2-11}
\cline{1-2}
\end{tabular}
    \caption{Results for the L63 system with $N=2\times10^4$, $\Delta t=0.02$ and $\varepsilon=\sqrt{0.05}\approx0.224$ for various surrogate models.}
    \label{tab:L63_0_s}
\end{table}


\subsubsection{Lorenz-96}
\label{ssec:ap-L96}

For the $40$-dimensional L96 system with forcing $F=10$ we use $(N, \Delta t, \varepsilon) = (10^5, 0.01, \sqrt{0.5})$ corresponding to the setup used in \cite{platt2022systematic, vlachas2020backpropagation}. Tables~\ref{tab:L96_1_s} and \ref{tab:L96_1_s-l} document the results for non-localized and localized architectures, respectively. Note that \cite{platt2022systematic} uses $F=8$ and \cite{vlachas2020backpropagation} uses both $F=8$ and $10$. Section~4 of \cite{vlachas2020backpropagation} shows that trained surrogate models demonstrate similar forecasting skill for both values of $F$. This justifies comparing our results with \cite{platt2022systematic, vlachas2020backpropagation}. To generate the training and testing data for the L96 system, we use a burn-in period of $1,000$ model time units. The training data matrix for the L96 system is well-conditioned, so introducing noise does not enhance the quality of the trained surrogate model (cf. the last row of Table~\ref{tab:L96_1_s-l} where we report results for LocalDeepSkip$_{2,2}$ trained on artificially noisy data).

\begin{table}[!htp]
    \centering
    \begin{tabular}{|c|c|c|c|c|c|c|c|c|c|c|} \hline
\multicolumn{4}{|c|}{Model} &\multicolumn{5}{c|}{VPT} & \multicolumn{2}{c|}{}\\ \hline
architecture & $D_r$ & $B$ & model size & mean & std & median & min & max &$\beta$ & $\mathbb{E}[t_{\rm train}]$(s)\\ \hline\hline
\multirow{5}{*}{SkipRFM} & 512 & 1 & 41,472 & 0.3 & 0.1 & 0.3 & 0.2 & 0.7 & 3.52e-09 & 1.6e-02\\ \cline{2-11}
 & 1,024 & 1 & 82,944 & 1.0 & 0.2 & 0.9 & 0.6 & 2.1 & 6.40e-09 & 2.4e-02\\ \cline{2-11}
 & 2,048 & 1 & 165,888 & 2.0 & 0.5 & 2.0 & 1.0 & 4.4 & 4.60e-08 & 6.6e-02\\ \cline{2-11}
 & 4,096 & 1 & 331,776 & 2.2 & 0.5 & 2.2 & 1.1 & 4.1 & 3.16e-07 & 2.3e-01\\ \cline{2-11}
 & \cellcolor{pink}8,192 & \cellcolor{pink}1 & \cellcolor{pink}663,552 & \cellcolor{pink}2.3 & \cellcolor{pink}0.6 & \cellcolor{pink}2.3 & \cellcolor{pink}1.2 & \cellcolor{pink}4.2 & \cellcolor{pink}3.16e-07 & \cellcolor{pink}1.0e+00\\ \cline{2-11}
\hline\hline
\multirow{5}{*}{DeepSkip} & 4,096 & 1 & 495,616 & 2.3 & 0.5 & 2.2 & 1.1 & 4.8 & 1.72e-07 & 2.4e-01\\ \cline{2-11}
 & 4,096 & 2 & 991,232 & 2.7 & 0.6 & 2.6 & 1.1 & 5.0 & 1.72e-07 & 5.0e-01\\ \cline{2-11}
 & 4,096 & 4 & 1,982,464 & 2.7 & 0.6 & 2.7 & 1.4 & 5.0 & 1.72e-07 & 1.0e+00\\ \cline{2-11}
 & 4,096 & 8 & 3,964,928 & 2.8 & 0.6 & 2.8 & 1.5 & 4.6 & 1.72e-07 & 2.0e+00\\ \cline{2-11}
 & \cellcolor{pink}4,096 & \cellcolor{pink}16 & \cellcolor{pink}7,929,856 & \cellcolor{pink}2.8 & \cellcolor{pink}0.6 & \cellcolor{pink}2.8 & \cellcolor{pink}1.5 & \cellcolor{pink}5.7 & \cellcolor{pink}1.72e-07 & \cellcolor{pink}4.0e+00\\ \cline{2-11}
\cline{1-2}
\end{tabular}
\caption{Results for non-localized architectures for the L96 system with $N=10^5$, $\Delta t=0.01$ and $\varepsilon=0.5$ for various surrogate models.}
    \label{tab:L96_1_s}
\end{table}

\begin{table}[!htp]
    \centering
    \begin{tabular}{|c|c|c|c|c|c|c|c|c|c|c|} \hline
\multicolumn{4}{|c|}{Model} &\multicolumn{5}{c|}{VPT} & \multicolumn{2}{c|}{}\\ \hline
architecture & $D_r$ & $B$ & model size & mean & std & median & min & max &$\beta$ & $\mathbb{E}[t_{\rm train}]$(s)\\ \hline\hline
\multirow{6}{*}{LocalSkip$_{2,2}$} & 512 & 1 & 6,656 & 4.4 & 0.9 & 4.3 & 2.0 & 7.6 & 3.16e-09 & 6.3e-02\\ \cline{2-11}
 & 1,024 & 1 & 13,312 & 5.3 & 1.1 & 5.3 & 1.9 & 9.1 & 3.16e-08 & 7.8e-02\\ \cline{2-11}
 & 2,048 & 1 & 26,624 & 5.7 & 1.0 & 5.7 & 3.2 & 8.9 & 8.92e-08 & 1.2e-01\\ \cline{2-11}
 & 4,096 & 1 & 53,248 & 6.5 & 1.2 & 6.4 & 3.5 & 11.1 & 1.00e-07 & 2.8e-01\\ \cline{2-11}
 & 8,192 & 1 & 106,496 & 6.7 & 1.1 & 6.8 & 4.1 & 10.4 & 4.24e-07 & 1.1e+00\\ \cline{2-11}
 & \cellcolor{pink}16,384 & \cellcolor{pink}1 & \cellcolor{pink}212,992 & \cellcolor{pink}6.8 & \cellcolor{pink}1.2 & \cellcolor{pink}6.8 & \cellcolor{pink}3.4 & \cellcolor{pink}10.7 & \cellcolor{pink}7.48e-07 & \cellcolor{pink}4.4e+00\\ \cline{2-11}
\hline\hline
\multirow{9}{*}{LocalDeepRFM$_{2,2}$} & 512 & 4 & 30,720 & 4.8 & 0.9 & 4.7 & 2.3 & 7.4 & 5.32e-09 & 4.7e-02\\ \cline{2-11}
 & 1,024 & 4 & 61,440 & 5.9 & 1.1 & 5.9 & 2.8 & 9.2 & 1.72e-08 & 9.1e-02\\ \cline{2-11}
 & 2,048 & 4 & 122,880 & 6.2 & 1.1 & 6.2 & 2.9 & 9.5 & 1.36e-07 & 2.7e-01\\ \cline{2-11}
 & 4,096 & 4 & 245,760 & 6.6 & 1.2 & 6.6 & 3.5 & 10.0 & 1.72e-07 & 9.4e-01\\ \cline{2-11}
 & 8,192 & 2 & 245,760 & 6.9 & 1.2 & 7.0 & 4.0 & 11.1 & 3.16e-07 & 2.0e+00\\ \cline{2-11}
 & 11,586 & 2 & 347,580 & 7.1 & 1.3 & 7.0 & 3.9 & 11.5 & 3.52e-07 & 4.3e+00\\ \cline{2-11}
 & 8,192 & 4 & 491,520 & 7.0 & 1.3 & 7.0 & 3.7 & 11.2 & 3.16e-07 & 4.2e+00\\ \cline{2-11}
 & \cellcolor{pink}16,384 & \cellcolor{pink}2 & \cellcolor{pink}491,520 & \cellcolor{pink}7.2 & \cellcolor{pink}1.3 & \cellcolor{pink}7.1 & \cellcolor{pink}3.9 & \cellcolor{pink}11.3 & \cellcolor{pink}3.88e-07 & \cellcolor{pink}9.5e+00\\ \cline{2-11}
 & 11,586 & 4 & 695,160 & 7.1 & 1.3 & 7.0 & 4.0 & 11.2 & 3.52e-07 & 8.8e+00\\ \cline{2-11}
\hline\hline
\multirow{1}{*}{LocalDeepSkip$_{1,4}$} & \cellcolor{pink}16,384 & \cellcolor{pink}2 & \cellcolor{pink}393,216 & \cellcolor{pink}6.9 & \cellcolor{pink}1.3 & \cellcolor{pink}6.9 & \cellcolor{pink}3.7 & \cellcolor{pink}11.1 & \cellcolor{pink}6.40e-07 & \cellcolor{pink}9.4e+00\\ \cline{2-11}
\hline\hline
\multirow{24}{*}{LocalDeepSkip$_{2,2}$} & 1,024 & 1 & 15,360 & 5.5 & 1.1 & 5.5 & 2.7 & 8.7 & 9.64e-09 & 9.7e-02\\ \cline{2-11}
 & 1,024 & 2 & 30,720 & 5.8 & 1.2 & 5.8 & 2.5 & 10.4 & 9.64e-09 & 1.2e-01\\ \cline{2-11}
 & 2,048 & 1 & 30,720 & 5.8 & 1.1 & 5.7 & 2.3 & 9.2 & 4.96e-08 & 1.3e-01\\ \cline{2-11}
 & 1,024 & 4 & 61,440 & 6.0 & 1.2 & 5.9 & 2.5 & 9.3 & 9.64e-09 & 1.6e-01\\ \cline{2-11}
 & 2,048 & 2 & 61,440 & 6.3 & 1.2 & 6.2 & 2.7 & 11.1 & 4.96e-08 & 2.0e-01\\ \cline{2-11}
 & 4,096 & 1 & 61,440 & 6.0 & 1.1 & 5.9 & 3.5 & 10.0 & 3.88e-07 & 3.2e-01\\ \cline{2-11}
 & 1,024 & 8 & 122,880 & 6.0 & 1.1 & 6.0 & 2.9 & 9.0 & 9.64e-09 & 2.5e-01\\ \cline{2-11}
 & 2,048 & 4 & 122,880 & 6.4 & 1.2 & 6.4 & 3.4 & 10.5 & 4.96e-08 & 3.4e-01\\ \cline{2-11}
 & 4,096 & 2 & 122,880 & 6.6 & 1.2 & 6.6 & 3.0 & 10.1 & 3.88e-07 & 5.8e-01\\ \cline{2-11}
 & 8,192 & 1 & 122,880 & 6.2 & 1.1 & 6.2 & 3.5 & 9.6 & 9.64e-07 & 1.2e+00\\ \cline{2-11}
 & 1,024 & 16 & 245,760 & 6.1 & 1.2 & 6.0 & 3.3 & 10.5 & 9.64e-09 & 4.3e-01\\ \cline{2-11}
 & 2,048 & 8 & 245,760 & 6.5 & 1.2 & 6.5 & 2.6 & 10.2 & 4.96e-08 & 6.1e-01\\ \cline{2-11}
 & 4,096 & 4 & 245,760 & 6.7 & 1.2 & 6.7 & 3.7 & 10.4 & 3.88e-07 & 1.1e+00\\ \cline{2-11}
 & 8,192 & 2 & 245,760 & 6.8 & 1.3 & 6.7 & 3.4 & 10.4 & 9.64e-07 & 2.3e+00\\ \cline{2-11}
 & 16,384 & 1 & 245,760 & 7.0 & 1.2 & 7.0 & 3.9 & 11.1 & 3.88e-07 & 4.5e+00\\ \cline{2-11}
 & 1,024 & 32 & 491,520 & 6.3 & 1.2 & 6.2 & 3.0 & 10.1 & 9.64e-09 & 7.8e-01\\ \cline{2-11}
 & 2,048 & 16 & 491,520 & 6.6 & 1.2 & 6.6 & 3.2 & 11.5 & 4.96e-08 & 1.1e+00\\ \cline{2-11}
 & 4,096 & 8 & 491,520 & 6.9 & 1.2 & 6.8 & 4.1 & 10.7 & 3.88e-07 & 2.1e+00\\ \cline{2-11}
 & 8,192 & 4 & 491,520 & 7.0 & 1.2 & 6.9 & 3.4 & 12.1 & 9.64e-07 & 4.5e+00\\ \cline{2-11}
 & \cellcolor{pink}16,384 & \cellcolor{pink}2 & \cellcolor{pink}491,520 & \cellcolor{pink}7.3 & \cellcolor{pink}1.2 & \cellcolor{pink}7.2 & \cellcolor{pink}4.3 & \cellcolor{pink}11.7 & \cellcolor{pink}3.88e-07 & \cellcolor{pink}9.1e+00\\ \cline{2-11}
 & 2,048 & 32 & 983,040 & 6.7 & 1.2 & 6.7 & 4.1 & 10.8 & 4.96e-08 & 2.2e+00\\ \cline{2-11}
 & 4,096 & 16 & 983,040 & 7.0 & 1.3 & 7.0 & 3.7 & 11.1 & 3.88e-07 & 4.1e+00\\ \cline{2-11}
 & 8,192 & 8 & 983,040 & 7.1 & 1.3 & 7.0 & 3.9 & 12.0 & 9.64e-07 & 8.9e+00\\ \cline{2-11}
 & 16,384 & 4 & 983,040 & 7.2 & 1.2 & 7.2 & 4.1 & 11.8 & 3.88e-07 & 1.8e+01\\ \cline{2-11}
\hline\hline
\multirow{1}{*}{LocalDeepSkipN$_{2,2}$} & \cellcolor{pink}16,384 & \cellcolor{pink}2 & \cellcolor{pink}491,520 & \cellcolor{pink}7.1 & \cellcolor{pink}1.3 & \cellcolor{pink}7.1 & \cellcolor{pink}3.8 & \cellcolor{pink}10.8 & \cellcolor{pink}3.88e-07 & \cellcolor{pink}9.5e+00\\ \cline{2-11}
\cline{1-2}
\end{tabular}
    \caption{Results for localized architectures for the L96 system with $N=10^5$, $\Delta t=0.01$ and $\varepsilon=0.5$ for various surrogate models.}
    \label{tab:L96_1_s-l}
\end{table}


\subsubsection{Kuramoto-Sivashinsky}
\label{ssec:ap-KS}

For the KS equation with domain length $L=200$ and $512$ spatial grid points we use $(N, \Delta t, \varepsilon) = (10^5, 0.25, 0.5)$, corresponding to the setup used in \cite{vlachas2020backpropagation}. To generate the training and testing data for KS, we use a burn-in period of $2.5\times10^4$ model time units. We used the following initial condition,
\begin{align}
    u(x,0)=\cos\left(\frac{2\pi x}{L}\right)\left(1+\sin\left(\frac{2\pi x}{L}\right)\right).
\end{align}
Table~\ref{tab:KS_1} documents the results. 

\begin{table}[!htp]
    \centering
    \begin{tabular}{|c|c|c|c|c|c|c|c|c|c|c|} \hline
\multicolumn{4}{|c|}{Model} &\multicolumn{5}{c|}{VPT} & \multicolumn{2}{c|}{}\\ \hline
architecture & $D_r$ & $B$ & model size & mean & std & median & min & max &$\beta$ & $\mathbb{E}[t_{\rm train}]$(s)\\ \hline\hline
\multirow{2}{*}{LocalRFM$_{8,1}$} & 8,192 & 1 & 270,336 & 3.6 & 1.3 & 3.9 & 0.5 & 6.5 & 8.56e-06 & 1.2e+00\\ \cline{2-11}
 & \cellcolor{pink}15,000 & \cellcolor{pink}1 & \cellcolor{pink}495,000 & \cellcolor{pink}4.3 & \cellcolor{pink}0.8 & \cellcolor{pink}4.3 & \cellcolor{pink}1.5 & \cellcolor{pink}6.3 & \cellcolor{pink}2.80e-05 & \cellcolor{pink}4.1e+00\\ \cline{2-11}
\hline\hline
\multirow{6}{*}{LocalDeepRFM$_{8,1}$} & 8,192 & 2 & 671,744 & 4.0 & 1.3 & 4.2 & 0.5 & 6.8 & 4.60e-06 & 2.1e+00\\ \cline{2-11}
 & 8,192 & 3 & 1,007,616 & 4.5 & 1.0 & 4.6 & 1.0 & 7.1 & 3.52e-05 & 3.1e+00\\ \cline{2-11}
 & \cellcolor{pink}14,000 & \cellcolor{pink}2 & \cellcolor{pink}1,148,000 & \cellcolor{pink}4.8 & \cellcolor{pink}1.0 & \cellcolor{pink}4.9 & \cellcolor{pink}2.1 & \cellcolor{pink}7.1 & \cellcolor{pink}2.00e-05 & \cellcolor{pink}6.5e+00\\ \cline{2-11}
 & 15,000 & 2 & 1,230,000 & 4.6 & 0.9 & 4.7 & 2.1 & 6.9 & 4.24e-05 & 7.4e+00\\ \cline{2-11}
 & 15,000 & 3 & 1,845,000 & 4.7 & 1.0 & 4.8 & 1.7 & 7.3 & 3.88e-05 & 1.1e+01\\ \cline{2-11}
 & 13,308 & 5 & 2,728,140 & 4.6 & 1.0 & 4.7 & 1.9 & 7.0 & 9.55e-05 & 1.5e+01\\ \cline{2-11}
\hline\hline
\multirow{1}{*}{LocalDeepSkip$_{8,1}$} & \cellcolor{pink}15,000 & \cellcolor{pink}2 & \cellcolor{pink}1,230,000 & \cellcolor{pink}0.5 & \cellcolor{pink}0.1 & \cellcolor{pink}0.5 & \cellcolor{pink}0.4 & \cellcolor{pink}0.8 & \cellcolor{pink}2.00e-05 & \cellcolor{pink}7.9e+00\\ \cline{2-11}
\hline\hline
\multirow{1}{*}{LocalRFMN$_{8,1}$} & \cellcolor{pink}15,000 & \cellcolor{pink}1 & \cellcolor{pink}495,000 & \cellcolor{pink}4.3 & \cellcolor{pink}0.9 & \cellcolor{pink}4.4 & \cellcolor{pink}2.0 & \cellcolor{pink}6.4 & \cellcolor{pink}4.24e-05 & \cellcolor{pink}3.8e+00\\ \cline{2-11}
\hline\hline
\multirow{1}{*}{LocalSkipN$_{8,1}$} & \cellcolor{pink}15,000 & \cellcolor{pink}1 & \cellcolor{pink}495,000 & \cellcolor{pink}4.3 & \cellcolor{pink}0.8 & \cellcolor{pink}4.4 & \cellcolor{pink}2.0 & \cellcolor{pink}6.4 & \cellcolor{pink}4.24e-05 & \cellcolor{pink}3.8e+00\\ \cline{2-11}
\hline\hline
\multirow{2}{*}{LocalDeepRFMN$_{8,1}$} & 14,000 & 2 & 1,148,000 & 4.9 & 0.9 & 5.1 & 2.7 & 7.0 & 2.00e-05 & 6.3e+00\\ \cline{2-11}
 & \cellcolor{pink}15,000 & \cellcolor{pink}2 & \cellcolor{pink}1,230,000 & \cellcolor{pink}5.0 & \cellcolor{pink}0.9 & \cellcolor{pink}5.0 & \cellcolor{pink}2.7 & \cellcolor{pink}7.6 & \cellcolor{pink}2.00e-05 & \cellcolor{pink}7.6e+00\\ \cline{2-11}
\hline\hline
\multirow{1}{*}{LocalDeepSkipN$_{8,1}$} & \cellcolor{pink}15,000 & \cellcolor{pink}2 & \cellcolor{pink}1,230,000 & \cellcolor{pink}5.0 & \cellcolor{pink}0.9 & \cellcolor{pink}5.1 & \cellcolor{pink}2.6 & \cellcolor{pink}7.7 & \cellcolor{pink}2.00e-05 & \cellcolor{pink}7.9e+00\\ \cline{2-11}
\hline\hline
\end{tabular}
    \caption{Results for the KS equation with $N=10^5$, $\Delta t=0.25$ and $\varepsilon=0.5$ for various surrogate models.}
    \label{tab:KS_1}
\end{table}


\subsection{Localization schemes}
\label{ssec:loc}

\begin{figure}[!htp]
    \centering
    \includegraphics[scale=0.55]{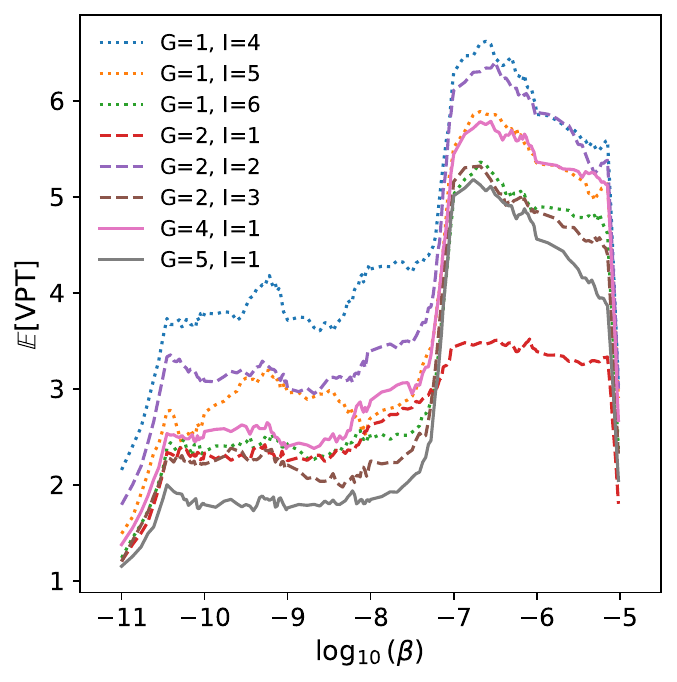}
    \caption{Estimates of the mean VPT as a function of the regularization hyperparameter $\beta$ for different localization schemes for the L96 system. The models depicted here are LocalSkip with $D_r=4,096$.}
    \label{fig:L96-loc}
\end{figure}

\begin{figure}[!htp]
    \centering
    \includegraphics[scale=0.4]{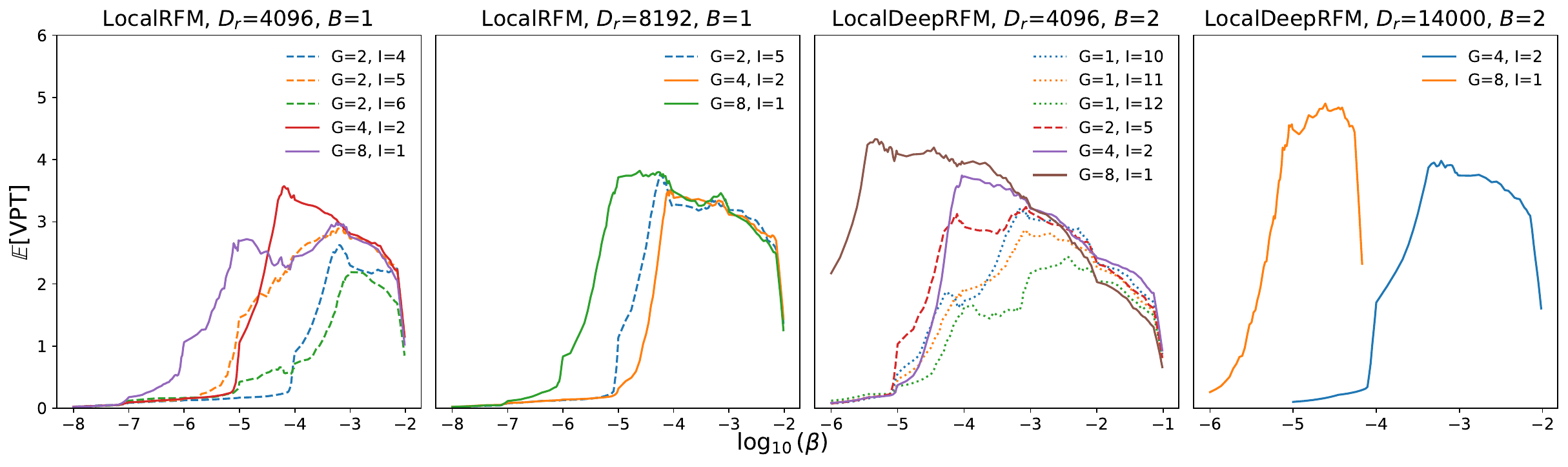}
    \caption{Estimates of the mean VPT as a function of the regularization hyperparameter $\beta$ for different localization schemes for the KS equation for various localized random feature models.}
    \label{fig:KS-loc}
\end{figure}

In this section we discuss the efficacy of various localization schemes for the $40$-dimensional  L96 system and the $512$-dimensional discretization of the KS equation. Figures~\ref{fig:L96-loc} and \ref{fig:KS-loc} show crude estimates of the mean VPT as a function of the regularization parameter $\beta$ for these systems, respectively. These estimates were computed by averaging over $5$ samples differing in the training data, the testing data and the non-trainable internal weights and biases for each value of $\beta$. The data shown in this section correspond to a fixed training data size $N=10^5$. Figure~\ref{fig:L96-loc} shows that $(G, I)=(1, 4)$ and $(2, 2)$ are the best performing localization schemes for L96 and Figure~\ref{fig:KS-loc} shows that overall $(G, I)=(8, 1)$ is the best performing localization scheme for KS. Comparing the first and second panels of Figure~\ref{fig:KS-loc} we see that the optimal localization scheme varies for different values of $D_r$.

While choosing a localization scheme, practitioners should consider several factors such as hardware e.g. available GPU memory, model size, $D_r$, amount of training data $N$, inferences drawn from the decay of the spatial correlation of the system, physical intuitions about the underlying dynamical system etc. If $G_2>G_1$ then for the same architecture and $D_r$, the model using scheme $(G_2, I_2)$ will typically have larger size compared to the model using scheme $(G_1, I_1)$. Since the GPU memory used during training is primarily a function of  $ND_r$,  for the same $N$ both models occupy roughly the same amount of memory on the GPU during training, despite the model using scheme $(G_1, I_1)$ having smaller size. Therefore, if our goal is to fit the largest possible model on the GPU during training, we should opt for the localization scheme with larger $G$. These considerations lead us to choose $(G, I)=(2, 2)$ for L96 and $(G, I)=(8, 1)$ for KS. These choices are consistent with those employed in \cite{platt2022systematic, vlachas2020backpropagation}.


\subsection{\textcolor{red-}{Interplay of width and depth in deep random feature models}}
\label{ssec:BDr}
\begin{figure}[!htp]
    \centering
\includegraphics[scale=0.5]{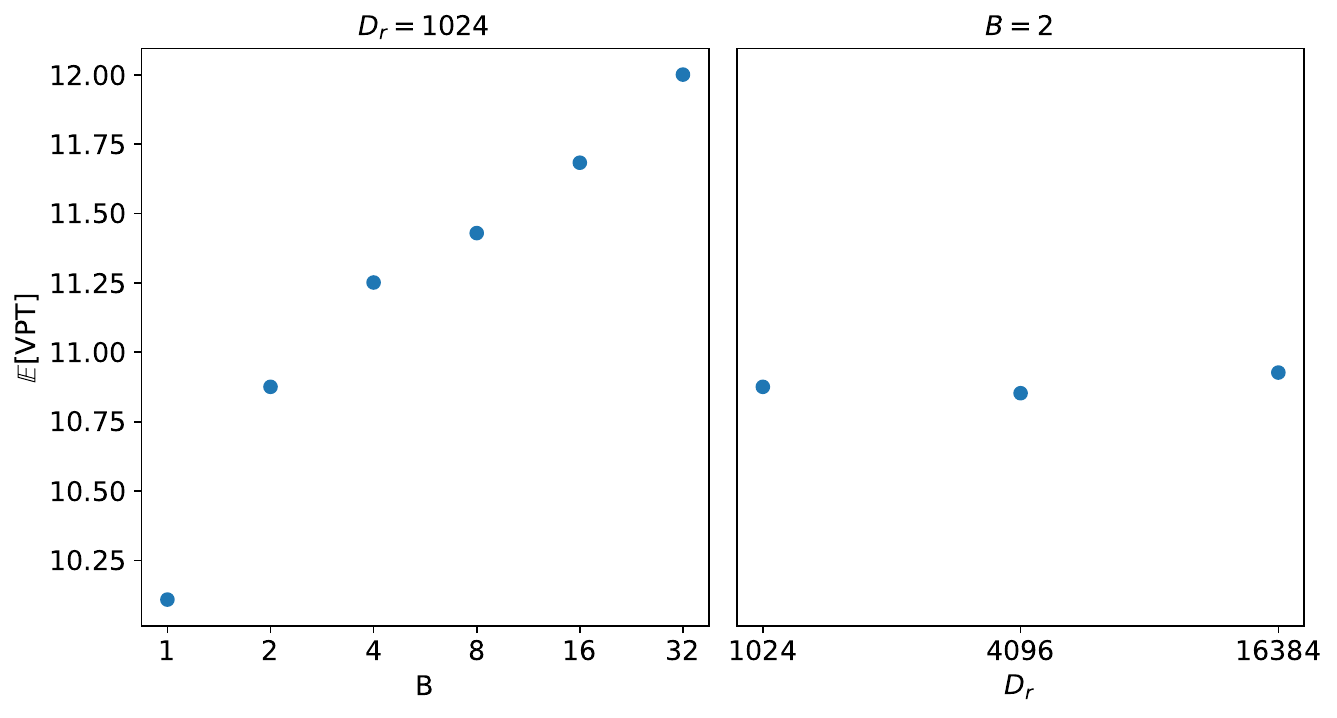}    
\caption{\textcolor{red-}{Mean VPT as a function of the depth $B$ with fixed width $D_r=1,024$ (left) and of the width $D_r$ with fixed depth $B=2$ (right) for the $3$-dimensional L63 system \eqref{eq:L63}. We show DeepSkip surrogate models with the data taken from Table~\ref{tab:L63_1_s}. Note that the overall model size $S$ increases with increasing depth (width).}}
    \label{fig:L63-ncs}
\end{figure}

\begin{figure}[!htp]
    \centering
\includegraphics[scale=0.5]{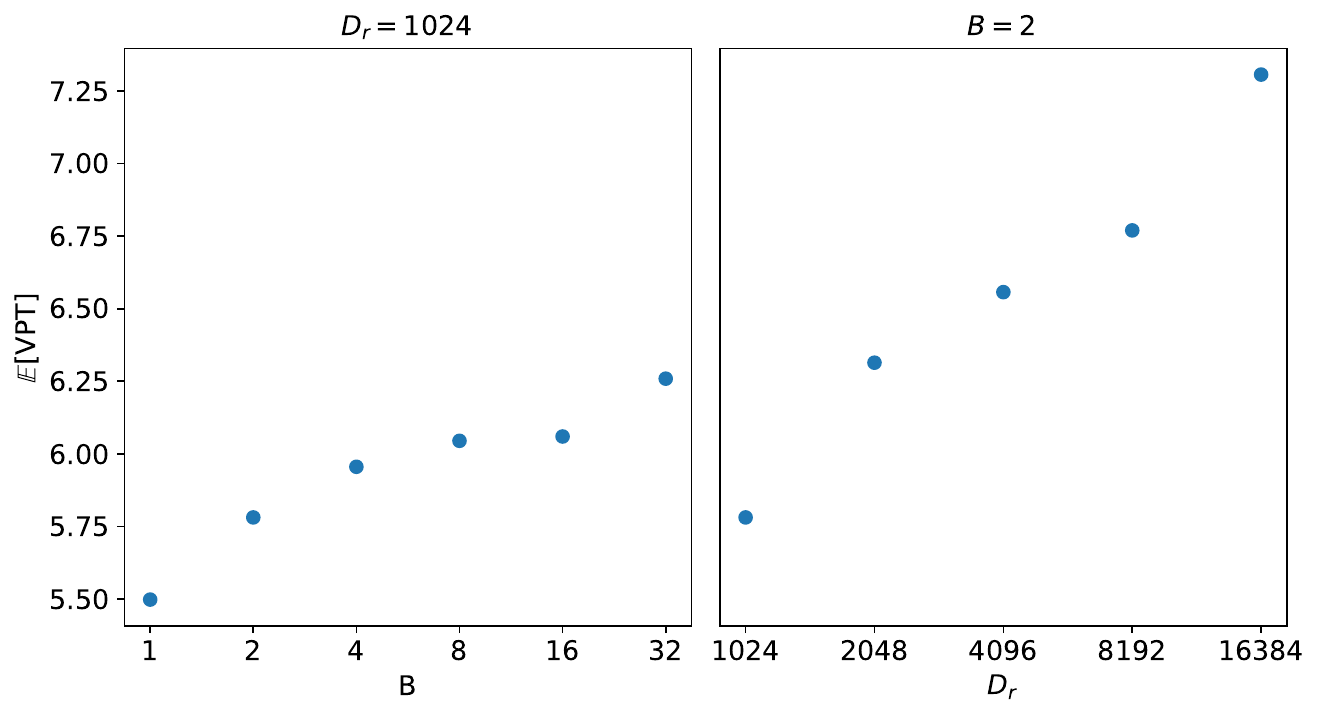}    
\caption{\textcolor{red-}{Mean VPT as a function of the depth $B$ with fixed width $D_r=1,024$ (left) and of the width $D_r$ with fixed depth $B=2$ (right) for the $40$-dimensional L96 system \eqref{eq:L63}. We show LocalDeepSkip$_{2,2}$ surrogate models with the data taken from Table~\ref{tab:L96_1_s-l}. Note that the overall model size $S$ increases with increasing depth (width).}}
\label{fig:L96-ncs}
\end{figure}

\textcolor{red-}{
In the main text we presented results for deep random feature architectures varying the depth $B$ while keeping the total model size constant, i.e. decreasing the width $D_r$. Whereas for the L63 system \eqref{eq:L63} the mean VPT increased with increasing depth, it decreased for the higher-dimensional L96 system \eqref{eq:L96} (cf. Figures~\ref{fig:L63-depth} and \ref{fig:L96-depth}, respectively). Here we show results where we vary the depth $B$ while keeping the width $D_r$ constant, and vice versa. Results for the L63 system are shown in Figure~\ref{fig:L63-ncs} and for the L96 system in Figure~\ref{fig:L96-ncs}. Figures~\ref{fig:L63-ncs} and \ref{fig:L96-ncs} confirm that increasing model size improves forecasting skill, until eventually saturation is reached due to the inherent chaotic nature of the dynamical system. As expected, increasing the depth $B$ while keeping the width $D_r$ constant leads to an increase of the mean VPT of roughly $20\%$ in both systems, when varying the depth from $B=2$ to $B=32$. This increase in forecast skill is consistent with the expectation that higher model sizes allow for better approximation. Varying the width is a bit more subtle. We see that for the small $3$-dimensional L63 system the forecasting skill has already saturated for the widths $D_r$ considered here with a mean VPT of $\sim 10.9$. In contrast, for the higher-dimensional L96 system the mean VPT does not saturate for the widths considered here and keeps increasing until we reach the maximal width $D_r$ compatible with the memory constraints of the GPU. Note that the increase in forecast skill is much stronger for varying the width $D_r$ than for varying the depth $B$. This explains the decrease of the forecast skill for the L96 system for increasing depth $B$ for a constant model size $S$ as seen in Figure~\ref{fig:L96-depth} and the increase of the forecast skill for the L63 system as seen in Figure~\ref{fig:L63-depth}. Hence, one should consider deep variants with $B>1$  once the width $D_r$ is sufficiently large such that saturation of the forecast skill has occurred or for designing robust surrogate models capable of handling larger sampling times (see Figure~\ref{fig:L63-large-dt}).} 

\subsection{Effect of depth on training time}
\label{ssec:time}

\begin{figure}[!htp]
    \centering
    \includegraphics[width=0.65\linewidth]{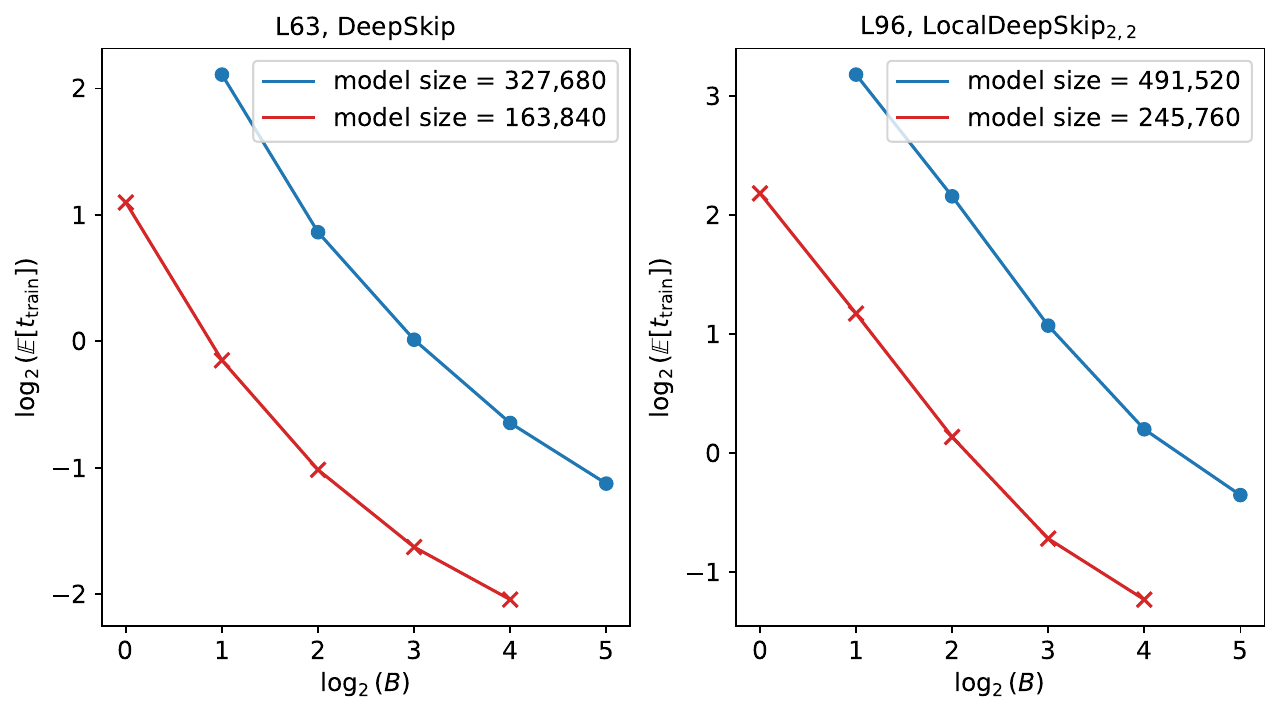}
    \caption{Average training time in seconds as a function of depth $B$. The left and right panels show results for DeepSkip and LocalDeepSkip taken from Tables~\ref{tab:L63_1_s} and \ref{tab:L96_1_s-l}, respectively. Along each curve the model size $S$ remains constant and the width $D_r$ decreases with depth.}
    \label{fig:time}
\end{figure}

In this section we demonstrate that deeper models train faster using the L63 system \eqref{eq:L63} as an example. Figure~\ref{fig:time} shows that for both non-localized and localized architectures, making a model deeper while keeping its model size $S$ fixed leads to faster training times. In fact, for both cases we see that the training time can be reduced by an order of magnitude by increasing the depth $B$. This is achieved because the linear regression problem occupies smaller space on the GPU for deeper models, as discussed in Section~\ref{ssec:deep}. 
Note that the training time shown here includes the run-time of the sampling algorithm~\ref{algo:hr}, which accounts for only a small fraction of the total time.

\end{document}